\documentclass[journal]{IEEEtran}
\usepackage{amsmath,amsfonts,amssymb,amsthm}
\usepackage{algorithmic}
\usepackage{algorithm}
\usepackage{array}
\usepackage[caption=false,font=normalsize,labelfont=sf,textfont=sf]{subfig}
\usepackage{textcomp}
\usepackage{stfloats}
\usepackage{cite}
\usepackage{url}
\usepackage{verbatim}
\usepackage{graphicx}
\usepackage[dvipsnames]{xcolor}
\usepackage{tikz}
\usepackage{marvosym}
\usepackage{bbm}
\usepackage{hyperref}
\usepackage{array,multirow}
\usepackage{cuted}

\preCutedStrip{%
  \noindent\smash{%
    \vrule width \columnwidth height 0.4pt
    \kern-0.4pt
    \vrule height 1ex
  }%
}
\postCutedStrip{%
  \noindent\smash{%
    \vrule width \columnwidth height 0pt
    \hspace{\columnsep}%
    \vrule height 1ex
    \kern-0.4pt
    \vrule width \columnwidth height 0.4pt
    \kern-\columnwidth
    \kern-\columnsep
  }%
}

\newtheorem{theorem}{Theorem}[section]

\newtheorem{definition}[theorem]{Definition}

\newcommand{\floor}[1]{\left\lfloor #1 \right\rfloor}

\DeclareMathOperator*{\argmax}{arg\,max}

\newcommand{\AM}[1]{\textcolor{black}{#1}}

\begin{document}

\usetikzlibrary{calc, arrows.meta}

\tikzset{
  pics/lightning/.style 2 args={code={
    \draw [thick, arrows={-Stealth[scale=1.2]}, red]
    (#1) -- ($(#1)!.5!(#2) - (.25,-.25)$) -- 
    ($(#1)!.5!(#2) - (-.25,.25)$) -- (#2);
  }}
}

\title{Graph Neural Network-Based Reinforcement Learning for Controlling Biological Networks --~the GATTACA Framework}

\iftrue
\author{Andrzej Mizera, Jakub Zarzycki\\
% \vspace{2em}\textbf{This work has been submitted to the IEEE for possible publication. Copyright may be transferred without notice, after which this version may no longer be accessible.}%
\thanks{A.~Mizera and J.~Zarzycki are with the Faculty of Mathematics, Informatics and Mechanics, University of Warsaw, Banacha 2, 02-097 Warsaw, Poland. A.~Mizera is with the IDEAS Research Institute, Królewska 27, 00-060 Warsaw, Poland, and J.~Zarzycki is with IDEAS NCBR Sp. z~o.o., Chmielna 69, 00-801 Warsaw, Poland. \emph{Corresponding author: A.~Mizera, \href{mailto:amizera@mimuw.edu.pl}{amizera@mimuw.edu.pl}.}}}
\else
\author{}
% \thanks{anonymous}
\fi



\hyphenation{CA-BE-AN}

\maketitle

\begin{abstract}
Cellular reprogramming, the artificial transformation of one cell type into another, has been attracting increasing research attention due to its therapeutic potential for complex diseases. However, identifying effective reprogramming strategies through classical wet-lab experiments is hindered by lengthy time commitments and high costs. 

In this study, we explore the use of deep reinforcement learning (DRL) to control Boolean network models of complex biological systems, such as gene regulatory and signalling pathway networks. We formulate a~novel control problem for Boolean network models under the asynchronous update mode, specifically in the context of cellular reprogramming. To solve it, we devise GATTACA, a~scalable computational framework. 

To facilitate scalability of our framework, we consider previously introduced concept of a~pseudo-attractor and improve the~procedure for effective identification of pseudo-attractor states. We then incorporate graph neural networks with graph convolution operations into the artificial neural network approximator of the DRL agent's action-value function. This allows us to leverage the available knowledge on the structure of a~biological system and to indirectly, yet effectively, encode the system's modelled dynamics into a~latent representation.
%To leverage the available knowledge on the structure of a~biological system and to indirectly, yet effectively, encode its modelled dynamics into a~latent representation, we incorporate graph neural networks with graph convolution operations into the artificial neural network approximator for the action-value function learned by the DRL agent. 

Experiments on several large-scale, real-world biological networks from the literature demonstrate the scalability and effectiveness of our approach.
\end{abstract}


\section{Introduction}
Complex diseases remain a~major challenge for biomedical research due to the intricate interactions between genes forming {\it gene regulatory networks} (GRNs). These networks exhibit complex dynamics emerging from the structure of the network: the state of the genome evolves in time, finally reaching some stable gene expression profiles characterising cell types or, more generally, stable cellular functional states~\cite{HEBI05}.

%referred to as \emph{attractors}. 
In some cases, when a~GRN is perturbed, it can reach an~``unhealthy'' state, that is otherwise inaccessible~\cite{Barabasi2011NetMed}. An~artificial change of the cell fate, referred to as {\it cellular reprogramming}, aims to solve this issue by steering the GRN back to a~``healthy'' state. However, identifying effective reprogramming targets and intervention strategies through wet-lab experiments alone is prohibitively expensive and time-consuming. This motivates the development of \emph{in-silico} methods. 

Capturing emergent properties of GRN dynamics requires the consideration of the system as a~whole. To this end, GRNs can be holistically modelled with the formalism of Boolean networks (BNs) originally introduced by Kauffman in~\cite{KS69}. Its simplicity enables tractable simulations and analysis of non-linear biological dynamics while retaining essential, emergent complex system behaviour. Cellular reprogramming can be formulated as a~control problem within the framework of BNs.

To support cellular reprogramming, control strategies must be computed for large-scale BNs, taking into account both the structure and dynamics of the network~\cite{GR16}. Scalability is the main concern when devising new computational methods for control of asynchronous BNs. Existing state-of-the-art methods and tools, such as CABEAN~\cite{SP21}, provide exact algorithms for attractor-based control. Nevertheless, they are limited in scalability because of the infamous ``curse of dimensionality'' problem, i.e. the size of the state space grows exponentially with the number of genes considered, which is handled by additional assumptions about network decomposability.

Deep reinforcement learning (DRL) has proven highly successful in decision problems characterised by very large state-action spaces~\cite{MuZero20}. Therefore, it seems a~promising alternative to traditional methods for controlling GRNs. In~\cite{ours_tcs}, pbn-STAC --~a~DRL-based tool for control of asynchronous BNs and probabilistic Boolean networks (PBNs) was introduced, which was a~result of exploration into the potential of DRL for developing scalable control methods for both BN and PBN models of GRNs in the context of cellular reprogramming. Therein, notions of a~pseudo-attractor and a~pseudo-attractor state were introduced, a~pseudo-attractor states identification procedure (PASIP) was devised, and a~novel control problem, i.e. the Source-Target Attractor Control, that corresponds with the problem of identifying cellular reprogramming strategies was formulated. The pbn-STAC framework was shown to be able to find optimal and sub-optimal control strategies between source and target attractors for BN and PBN models of sizes up to 33 nodes.

In this study, we build on the initial idea of~\cite{ours_tcs} to devise a~general and scalable approach capable of handling very large BN models of biological networks. To this end, we introduce the \underline{G}raph-based \underline{At}tractor-\underline{Ta}rget \underline{C}ontrol \underline{A}lgorithm (GATTACA) computational framework to solve a~novel, general target-control problem for BN models of biological networks under the asynchronous update mode. Our new control problem is tailored for cellular reprogramming in the sense that control actions are allowed to take place only in attractor states as they correspond to observable phenotypic cellular states understood as discrete, mutually exclusive, stable, and observable cell phenotypes or cell fates that can undergo transitions from one to another under the influence of regulatory inputs. This approach addresses the constraints of wet-lab biology, where it is very difficult to experimentally determine a~transient state of a~cell, without destroying it, for the application of a~control intervention in a~real-world scenario. In contrast, stable cell phenotypes, which correspond to the attractor states of a~dynamical system~\cite{Grunt25}, are expected to be more easily, optically observable realisations of complex interactions between the genome, epigenome, and local environment. To reflect these constraints, we restrict control actions to be applied only in observable phenotypic cellular states, which are represented by attractors of a~BN model.

The core of our GATTACA framework is a~Graph Neural Network (GNN)-based Q-learning agent which learns minimal control strategies that drive the system from a~given source attractor state to one aligned with a~given target configuration, i.e. an~attractor with user-specified values for a~subset of nodes. The structure of the biological network is encoded directly into the GNN, allowing the model to exploit topological relationships between the network components (nodes) during training. Furthermore, to facilitate effective identification of attractors in large BN models of complex networks, we consider the notion of pseudo-attractor states introduced in~\cite{ours_tcs} and we enhance the pseudo-attractor states identification procedure proposed therein.

To assess the scalability potential, we evaluate the GATTACA framework on several large-scale, real-world biological networks, demonstrating its scalability and effectiveness in cases where exact symbolic approaches, such as those implemented in CABEAN, fail. Our contributions result in a~scalable and effective computational method for cellular reprogramming scenarios.
%Our contributions represent a~step forward in developing predictive, scalable and effective computational methods for cellular reprogramming scenarios.

The paper is organised as follows. Related works are discussed in Section~\ref{sec:related}. The preliminaries are presented in Section~\ref{sec:prelim}. Our novel target control problem in the context of cellular reprogramming is formulated in Section~\ref{sec:control_problem}. Next, a~DRL-based framework for controlling large-scale biological networks is developed in Section~\ref{sec:algorithm}. The framework is evaluated on large models of real-world gene regulatory and signalling networks in accordance with a~methodology presented in Section~\ref{sec:experiments}. The obtained evaluation results are discussed in Section~\ref{sec:results}. Finally, the study is concluded in Section~\ref{sec:discussions}.

\section{Related works}
\label{sec:related}

The existing structure- and dynamics-based state-of-the-art computational techniques are limited to small and mid-size networks due to the infamous state space explosion problem. They usually require the systems to facilitate some kind of decomposition. However, this is often too restrictive for cellular reprogramming considerations.

% Batch Reinforcement Learning
The application of reinforcement learning to control GRNs was pioneered in~\cite{SPA13}. The study focused on avoiding undesirable states in terms of steady-state probabilities of PBNs. The authors treated time-series gene expression samples as a~sequence of experience tuples and used a~batch version of Q-Learning to approximate the optimal policy. Later, to relax the temporal condition, the BOAFQI-Sarsa method was devised in~\cite{NCB18}. A~further development was mSFQI method proposed in~\cite{NBC20} for control based on the probabilities of gene activity profiles.

Rule-based reinforcement learning in the form of an~eXtended Classifier System (XCS) was devised in~\cite{Karlsen2018} for the control of random BNs. Subsequently, it was applied in~\cite{Karlsen2019} to a~11-node yeast cell cycle network. % from~\cite{yeast}.

% Deep Reinforcement Learning
Scalability is the main concern when devising new methods for control of BNs. This issue can be addressed by devising new methods based on deep reinforcement learning (DRL) techniques, which have proven highly successful in decision problems characterized by large state-action spaces, see, e.g.~\cite{MuZero20}. To this end, the idea of combining Q-learning with deep learning was conceived. The first to apply this idea in the context of BN control were Papagiannis \emph{et al.} in~\cite{Papagiannis2019}, where a~control method for BNs based on deep Q-Learning with experience reply, namely Double Deep Q-Network (DDQN) with Proritised Experience Replay (PER), was proposed. % initialy introduced in~\cite{ddqn}.
It was applied in~\cite{Papagiannis2021} to control a~20-node synthetic PBN and a~7-node real-world PBN model of melanoma. The same approach was used by Acernese \emph{et al.} in~\cite{Acernese2020} for the control of synchronous PBNs with the goal of driving the networks from a~particular state towards a~more desirable one. Next, in~\cite{Acernese2021}, the authors proposed a~Q-learning based method to address the feedback stabilisation problem of probabilistic Boolean control networks (PBCNs) with the aim of stabilising the system at a~given equilibrium point and applied it to a~small 9-node model of apoptosis. 

To address the control problem on large networks, Moschoyiannis \emph{et al.} in~\cite{Moschoyiannis2023} applied DDQN with PER to control a~large synchronous melanoma PBN model of 200 nodes. % To make their method scalable
They limited the control actions to flipping a~single specific gene, namely \emph{pirin}, which had already been identified as relevant for the control of melanoma PBN models in previous studies. Furthermore, the target set of states consisted of exactly half of the entire state space, that is, the objective was to reach a~subset of network states where the WNT5A gene is OFF.

Finally, Mizera \emph{et al.} in~\cite{ours_tcs} leveraged DRL to develop pbn-STAC, a~computational framework that solves the source-target attractor control problem of driving the network dynamics from a~given source to a~specified target attractor of a~BN or PBN model under the asynchronous update mode. The control problem is tailored for cellular reprogramming in the sense that control actions are allowed to take place only in intermediate attractor states to reflect the constraints of wet-lab biology. The approach was successful in controlling BN and PBN models of up to 33-nodes.

\section{Preliminaries}
\label{sec:prelim}
\subsection{Boolean networks}
A~\emph{Boolean function} $f$ is a~function of $n \in \mathbb{N}$ Boolean variables, $f : \{0,1\}^k \to \{0,1\}$, whose arguments and output take values in the binary set $\{0,1\}$. A~variable $x_l$ is an~\emph{essential variable} of a~Boolean function $f$ if and only if $f(x_1, \ldots, x_{l-1}, 0, x_{l+1}, \ldots, x_n) \neq f(x_1, \ldots, x_{l-1}, 1, x_{l+1}, \ldots, x_n)$ for some fixed values of $x_1, \ldots, x_{l-1}, x_{l+1}, \ldots, x_n$. A~\emph{Boolean network} $B$ is a~discrete dynamical system defined as a~tuple $B = (\boldsymbol{x}, \boldsymbol{f}),$ where $\boldsymbol{x}=(x_1,\allowbreak x_2,\ldots,x_n)$ is a~vector of binary-valued variables (also referred to as nodes or genes)
%, i.e. $x_i \in \{0,1\}$ for all $i=1,2\ldots,n$,
and $\boldsymbol{f}=(f_1,\allowbreak f_2, \ldots, f_n)$ is a~vector of Boolean functions. Each $x_i of \textbf{x}$ is associated with a~\emph{predictor function} (or simply a~\emph{predictor}) $f_i(x_{i_1}, x_{i_2}, \ldots, x_{i_{l(i)}}),$ which is the $i$-th component of $\boldsymbol{f}$ and is defined with respect to $1 \leq l(i) \leq n$ distinct variables from $\boldsymbol{x}$. The set of variables $\{x_{i_1}, x_{i_2}, \ldots, x_{i_{l(i)}}\}$ is referred to as the set of \emph{parent nodes} of $x_i$, denoted $\textrm{Pa}(x_i)$. We will assume that parent nodes $\textrm{Pa}(x_i)$ correspond to the set of essential variables of $f_i$. The predictor function $f_i$ defines the evolution of variable $x_i$ in discrete time steps.

The \emph{structure graph} of $B = (\textbf{x}, \textbf{f})$ is the directed graph $SG_{\textrm{B}} = (V,E)$, where $V$ is the set of nodes consisting of the variables of $\textbf{x}$ and $E = \{(x_j,x_i) \in V \times V \mid x_j \in \textrm{Pa}(x_i)\}$. The structure graph represents the individual direct regulatory relationships between the components of the system encoded by the respective Boolean functions.

The system dynamics evolves iteratively, with the states of the nodes being updated in accordance with their respective predictor functions. Two update modes are usually considered: \emph{synchronous}, where all nodes are updated at once, and \emph{asynchronous}, where one randomly chosen node is updated at a~time. As in~\cite{ours_tcs}, also in this study we focus on the asynchronous mode, which is considered more appropriate for the modelling of biological networks~\cite{CFM20,ZYLW+13}, yet more challenging from a~computational point of view.

The \emph{state space} of a~BN with $n$ nodes is $S=\{0,1\}^n$. At each time point $t$, the \emph{state} of the network is defined by the vector $\boldsymbol{x}(t)=(x_1(t),x_2(t),\ldots,x_n(t))$, where $x_i(t)$ is the value of variable $x_i$ at time $t$, i.e. $x_i(t) \in \{0,1\}$. Under the asynchronous update mode, a~single randomly selected node is updated at each time step according to its corresponding predictor function. The~associated asynchronous \emph{state transition graph} (STG) of $B$, denoted $\mathrm{STG}(B)$, is a~directed graph that represents the possible state transitions of $B$ under the asynchronous update mode. Formally, $\mathrm{STG}(B) = (S,\rightarrow)$, where $S = \{0,1\}^n$ is the set of all possible states and $\rightarrow$ is the set of directed edges such that a~directed edge from $\mathbf{s}=(s_1,s_2,\ldots,s_n) \in S$ to $\mathbf{s'}=(s_1',s_2',\ldots,s_n') \in S$, denoted $\mathbf{s} \rightarrow \mathbf{s'}$, is in $\rightarrow$ if and only if $\mathbf{s'}$ can be obtained from $\mathbf{s}$ by a~single asynchronous update, i.e. there exists $i = 1, 2, \ldots, n$, such that
\[
\begin{array}{lcl}
\left(s_i' = f_i(s_{i_1},s_{i_2},\cdots,s_{i_{l(i)}})\right)
\wedge \left(\bigwedge_{j\neq i}\ (s_j' = s_j)\right).
\end{array}
\]

For a~BN $B$ under the asynchronous update mode, a~path from a~state $\mathbf{s}$ to a~state $\mathbf{s'}$ is a~(possibly empty) sequence of transitions from $\mathbf{s}$ to $\mathbf{s'}$ in $\mathrm{STG}(B)$. A~path from a~state $\mathbf{s}$ to a~subset $S'$ of $S$ is a~path from $\mathbf{s}$ to any state $\mathbf{s'} \in S'$. For any state $\mathbf{s} \in S$, let $\mathsf{pre}(\mathbf{s}) = \{\mathbf{s'} \in S \mid \mathbf{s'} \rightarrow \mathbf{s}\}$ and let $\mathsf{post}(\mathbf{s}) = \{\mathbf{s'} \in S \mid \mathbf{s} \rightarrow \mathbf{s'}\}$. $\mathsf{pre}(\mathbf{s})$ contains all the states that can reach $\mathbf{s}$ by performing a~single transition in $\mathrm{STG}(B)$ and $\mathsf{post}(\mathbf{s})$ contains all the states that can be reached from $\mathbf{s}$ by a~single transition in $\mathrm{STG}(B)$. The definition of $\mathsf{post}$ can be lifted to a~subset $S' \subseteq S$ as:
\[
\mathsf{post}(S') = \bigcup_\mathbf{s \in S'}(\mathsf{post}(\mathbf{s})).
\]

For a~state $\mathbf{s} \in S$, $\mathsf{reach}(\mathbf{s})$ denotes the set of states $\mathbf{s'}$ such that there is a~path from $\mathbf{s}$ to $\mathbf{s'}$ in $\mathrm{STG}(B)$ and can be defined as the transitive closure of the $\mathsf{post}$ operation, i.e. we define:
\begin{align*}
\mathsf{post}^0(\mathbf{s}) &= \{\mathbf{s}\},\\
\mathsf{post}^{i+1}(\mathbf{s}) &= \mathsf{post}(\mathsf{post}^i(\mathbf{s})),
\end{align*}
and
\[
\mathsf{reach}(\mathbf{s}) = \bigcup_{i=0}^\infty(\mathsf{post}^i(\mathbf{s})).
\]
Thus, $\mathsf{reach}(\mathbf{s})$ is the smallest subset of states in $S$ such that $\mathbf{s} \in \mathsf{reach}(\mathbf{s})$ and $\mathsf{post}(\mathsf{reach}(\mathbf{s})) \subseteq \mathsf{reach}(\mathbf{s})$.

An~\emph{attractor} $A$ of $\mathrm{STG}(B)$ is a~subset of states of $S$ such that for every $\mathbf{s} \in A$, $\mathsf{reach}(\mathbf{s}) = A$. It is a~bottom strongly connected component (BSCC) in the STG of the network. A~\emph{fixed-point attractor} and a~\emph{multi-state attractor} are BSCCs consisting of a~single state and more than one state, respectively.

In the case of the asynchronous update mode, the dynamics of a~BN is non-deterministic, i.e. $\rightarrow$ is a~relation rather than a~function. It can be modelled as a~Markov chain. In this view, an~attractor of the BN corresponds to an~irreducible subset of states of the Markov chain.

For some examples of the introduced concepts related to BNs, we refer to Appendix~\ref{app:bn_example}.

\subsection{Pseudo-attractors}
It was shown that obtaining the family of all attractors of a~BN is NP-hard by itself~\cite{Akutsu2003}. This is especially prohibitive in the case of cellular reprogramming considerations, where ultimately large models need to be analysed. However, from a~practical point of view in the context of cellular reprogramming, only the frequently revisited states of an~attractor are the relevant ones since they correspond to phenotypical cellular states that are observable in the lab. This makes these states `recognisable' for the application of reprogramming interventions in practice in accordance with the control strategy obtained with our framework. With this in mind, the notions of a~\emph{pseudo-attractor} and a~\emph{pseudo-attractor state} were introduced in~\cite{ours_tcs}. Since in this study our focus is solely on BN models under the asynchronous update mode, herein we provide a~variant of the definition explicitly referring to asynchronous BNs.

\begin{definition}[Pseudo-attractor]
\label{def:pseudo-attractor}
Let $\mathcal{A}$ be an~attractor of an~asynchronous BN, i.e. an~irreducible set of states of the Markov chain underlying the BN. Let $n := |\mathcal{A}|$ be the size of the attractor $\mathcal{A}$ and let $\mathbb{P}_\mathcal{A}$ be the unique stationary probability distribution on $\mathcal{A}$. The \emph{pseudo-attractor} associated with $\mathcal{A}$ is the maximal subset $\mathcal{PA} \subseteq \mathcal{A}$ such that for all $s \in \mathcal{PA}$ it holds that $\mathbb{P}_\mathcal{A}(s) \geq \frac{1}{n}$. The states of a~pseudo attractor are referred to as \emph{pseudo-attractor states}.
\end{definition}

Notice that in accordance with the definition, a~pseudo-attractor is a~subset of an~attractor and pseudo-attractor states are attractor states. 

The correctness of the definition is guaranteed by the fact that the state space of the underlying Markov chain of an~asynchronous BN is finite and that the Markov chain restricted to the attractor states is irreducible. It is a~well known fact that all states of a~finite and irreducible Markov chain are positive recurrent. In consequence, the attractor-restricted Markov chain has a~unique stationary distribution.

In \cite{ours_tcs} it was proved that for any BN attractor there exists a~non-empty pseudo-attractor as stated by the following two theorems.

\begin{theorem}
\label{th:existence}
Let $\mathcal{A}$ be an~attractor of a~BN. Then there exists a~pseudo-attractor $\mathcal{PA} \subseteq \mathcal{A}$ such that $|\mathcal{PA}| \geq 1$.     
\end{theorem}

\begin{theorem}
\label{th:uniform}
%In the case of the~uniform stationary distribution on an~attractor, the associated pseudo-attractor is equal to the attractor:
Let $\mathcal{A}$ be an~attractor of a~BN such that the unique stationary distribution of the underlying Markov chain of the BN restricted to $\mathcal{A}$ is uniform. Then, for the pseudo-attractor $\mathcal{PA}$ associated with $\mathcal{A}$ it holds that $\mathcal{PA}=\mathcal{A}$, i.e. the pseudo-attractor and the attractor are equal.
\end{theorem}

Moreover, we proposed the pseudo-attractor state identification procedure (PASIP) that facilitates the determination of PA states in large models. In this study, we identify a~condition which may result in the blockage of the original PASIP. We provide a~solution to this issue and present here the improved version of PASIP. The details of the method are presented and discussed in Section~\ref{sec:pasip}.

\subsection{Reinforcement learning}
A~control problem for BNs can be viewed as a~Markov Decision Process (MDP) and can be solved using reinforcement learning (RL). RL solves sequential decision problems by optimizing a~cumulative reward function.

The RL problem is meant to be a~direct framing of the task of learning from interactions to achieve a~goal. The decision-maker, or learner, is referred to as an~\emph{agent}. At each discrete time step $t$, the agent interacts with an~\emph{environment}, which encompasses everything external to the agent. The agent's behaviour is defined by a~\emph{policy}, given by a~mapping $\pi:\mathcal{S} \times \mathcal{AC} \rightarrow [0,1]$ defined for all observable states in the environment $\mathcal{S}$ and all possible actions in action space $\mathcal{AC}$, where $\pi(a \mid s)$ is the probability of selecting action $a \in \mathcal{AC}$ while being in state $s \in \mathcal{S}$. This policy guides the agent's decision-making process throughout its interaction with the environment.

Within the MDP framework, the RL problem is formalised as the optimal control of an~MDP. RL methods specify how the agent changes its policy as a~result of its experiences with the goal of maximising the expected \emph{discounted return}:
\[
G_t \triangleq \sum_{k=0}^{\infty} \gamma^k R_{t+k+1},
\]
where $0 \leq \gamma \leq 1$ is the \emph{discount factor}.

Nearly all RL algorithms involve estimating \emph{value functions}. Herein, we focus on the \emph{action-value function} $q_\pi(s,a)$ defined as the expected return starting from $s$, taking the action $a$, and thereafter following policy $\pi$:
\begin{align*}
q_\pi(s,a) &\triangleq % \mathbb{E}_\pi[G_t \mid S_t = s, A_t = a]\\ &=
\mathbb{E}_\pi\left[ \sum_{k=0}^{\infty} \gamma^k R_{t+k+1} \biggm\vert S_t = s, A_t = a\right]\!.
\end{align*}
The value function satisfies a~particular recursive relationship expressed in the form of a~\emph{Bellman equation}, i.e.
\begin{multline}
\label{eq:bellman}
q_\pi(s,a) = \sum_{s' \in \mathcal{S}} P_a(s,s') \Bigl [ R_a(s,s')\\ + \gamma \sum_{a' \in \mathcal{AC}} \pi (s',a') q_\pi(s',a') \Bigr].
\end{multline}

Value functions define a~partial order over policies and there always exists at least one policy which is better than or equal to all other policies~\cite{SB18}. This policy, denoted $\pi_*$, is referred to as the~\emph{optimal policy}. Optimal policies share the same \emph{optimal action-value function}, denoted $q_*$, which is defined as
\[
q_*(s,a) \triangleq \max_\pi q_\pi(s,a),
\]
for all $s \in \mathcal{S}$ and $a \in \mathcal{AC}$. It holds that
\[
q_*(s,a) = \mathbb{E}_{\pi_*} \bigl [R_{t+1} + \gamma v_*(S_{t+1}) \bigm \vert S_t = a, A_t = a \bigr ].
\]

The optimal action-value function satisfies the \emph{Bellman optimality equation}:
\begin{align*}
q_*(s,a) &= \mathbb{E} \bigl [ R_{t+1} + \gamma\max_{a' \in \mathcal{AC}} q_*(S_{t+1},a') \bigm \vert S_t= s, A_t = a\bigr]\\
&= \sum_{s' \in \mathcal{S}} P_a(s,s') \bigl [R_a(s,s') + \gamma \max_{a' \in \mathcal{AC}} q_*(s',a') \bigr ],
\end{align*}
where the expectations are taken without reference to any specific policy since the value functions are optimal.

Optimal policies can be computed with \emph{dynamic programming} (DP) algorithms, such as \emph{policy iteration} or \emph{value iteration}, given a~perfect model of the environment in the form of an~MDP. DP algorithms are obtained by turning Bellman equations into update rules that improve the approximations of the desired value functions. However, the utility of DP algorithms is limited due to the assumption of a~perfect model and its computational expense. Therefore, for large state space environments or cases where the model of the environment is unavailable, Monte Carlo methods or temporal-difference learning (TD) methods are used, see~\cite{SB18}.

\subsection{Q-learning}
\label{sec:q-learning}

Q-learning is a~TD control algorithm and is one of the most popular model-free methods to solve RL problems. It is used to learn the optimal action-selection policy. The \emph{one-step Q-learning}
%, which is the simplest form of Q-learning,
is defined by the update rule 
\begin{align*}
Q(S_t, A_t) \leftarrow Q(S_t,A_t)+\alpha[\,R_{t+1}+\gamma\max_{a \in \mathcal{AC}} Q(S_{t+1},a)\\-Q(S_t,A_t)\!],
\end{align*}
where $0 < \alpha \leq 1$ is the \emph{learning rate}, which determines how fast past experiences are forgotten by the agent, and the learned action-value function $Q$ directly approximates $q_*$. The algorithm can be outlined as follows. First, initialise $Q(s,a)$ for all $(s,a) \in \mathcal{S} \times \mathcal{AC}$ and start the environment at state $S_0$. For time steps $t = 0, 1, 2, \ldots$, choose action $A_t$ according to the \emph{$\epsilon$-greedy policy}, i.e.
\[
A_t \leftarrow 
\begin{cases}
\argmax_{a \in \mathcal{AC}} Q(S_t,a) & \textrm{with prob. } 1 - \epsilon\\
\textrm{uniformly random action in } \mathcal{AC} & \textrm{with prob. } \epsilon,\\
\end{cases}
\]
and apply action $A_t$ in the environment. As a~result, the state is changed from $S_t$ to $S_{t+1} \sim P_a(s,\cdot)$. Observe $S_{t+1}$ and $R_{t+1}$. Finally, revise the action-value function $Q$ at state-action $(S_t, A_t)$ in accordance with the above update rule.

All that is required to assure convergence to $q_*$ is that all $(s,a) \in \mathcal{S} \times \mathcal{AC}$ pairs continue to be updated. Notice that the convergence is guaranteed independently of the policy being followed. This property renders Q-learning an~\emph{off-policy} TD control algorithm and enables the use of the $\epsilon$-greedy policy, which is a~simple technique for handling the exploration-exploitation tradeoff, to ensure the convergence.

%%% ====================================================================================================
%%% Uncomment for longer version 
%%% ====================================================================================================

%The Q-learning algorithm iteratively updates the Q-values based on observed transitions, gradually refining its estimates of the optimal action-value function. All that is required to assure convergence to $q_*$ is that all $(s,a) \in \mathcal{S} \times \mathcal{A}$ pairs continue to be updated. Notice that the convergence is guaranteed independently of the policy being followed. This property renders Q-learning an~\emph{off-policy} TD control algorithm and enables the use of the $\epsilon$-greedy policy, which is a~simple technique for handling the exploration-exploitation tradeoff, to ensure the convergence. Once learned, the Q-values are used to provide the agent with the optimal policy for selecting actions in each state.

%%% ====================================================================================================

\subsection{Deep reinforcement learning}
\label{sec:Q-function_approx}

The Q-learning algorithm has proven to be successful in controlling small-scale environments. However, in the case of large state-action spaces, Q-learning cannot learn all action values in all states separately and can fail to converge to an~optimal solution. To address this issue, the idea was conceived to learn a~parameterised value function $Q(s,a; \theta)$ by updating the parameters $\theta$ in accordance with the standard Q-learning-derived rule:
\begin{multline*}
\theta_{t+1} \leftarrow 
\theta_{t} + \alpha \Big[R_{t+1} + \gamma \max_{a \in \mathcal{AC}} Q(S_{t+1}, a; \theta_t)\\
-  Q(S_t, A_t; \theta_t) \Big] \nabla_{\theta_t} Q(S_t, A_t; \theta_t),
\end{multline*}
resembling stochastic gradient descent. It was shown that as the agent explores the environment, the parameterized value function converges to the true $Q$-function, see, e.g.~\cite{SB18}. 

The above idea is realized in the deep RL (DRL) framework, which combines classical RL with artificial neural network (ANN) function approximation. In DRL, a~function approximator is trained to estimate the Q-values. The approximator is a~deep Q-network agent (DQN), i.e. a~multi-layer ANN which for a~given observed state of the environment $s$ outputs a~vector of action values $Q(s,\cdot; \theta)$, where $\theta$ are the parameters of DQN. The aim of DQN is to approximate $q_*$ by iteratively minimising a~sequence of loss functions $\mathcal{L}_t(\theta_t)$:
\begin{align*}
\mathcal{L}_t(\theta_t) = \left( Y_t^{\textrm{DQN}} - Q(S_t, A_t; \theta_t) \right)^2,
\end{align*}
where $Y_t^{\textrm{DQN}}$ is the DQN target defined as
\begin{align*}
Y_t^{\textrm{DQN}} \triangleq R_{t+1} + \gamma \max_{a \in \mathcal{AC}} Q(S_{t+1}, a; \theta_t^-),
\end{align*}
while $\theta_t^-$ and $\theta_t$ are the parameters of the target and online networks, respectively. The target network is the same as the online network, but its parameter values are copied from the online network every $k$ steps and kept fixed at other steps.

\subsubsection{Double Deep Q-Network (DDQN)}
\label{sec:ddqn}

Both standard Q-learning and DQN use the same values for both selecting and evaluating actions, which can lead to choosing actions based on overestimated values~\cite{HGS16}. To address this issue, DDQN introduces two networks with separate sets of weights, $\theta$ and $\theta'$, where the former is used to predict actions and the latter to evaluate the greedy policy. They are trained by assigning each experience randomly to update one of the networks. DDQN uses the following DDQN target:
\begin{align*}
Y_t^{\textrm{DDQN}} \triangleq R_{t+1} + \gamma Q(S_{t+1}, \argmax_{a \in \mathcal{AC}} Q(S_{t+1}, a; \theta_t);\theta_t').
\end{align*}

\subsubsection{Branching Dueling Q-Network (BDQ)}
\label{sec:bdq}

BDQs are designed to address complex and high-dimensional action spaces; they enhance the scalability and sample efficiency of RL algorithms in complex scenarios. BDQs can be thought of as an~adaptation of the dueling network architecture~\cite{WSHH+16} into the action branching architecture. A~BDQ uses two separate ANNs, i.e. the \emph{target network} $Q^{-}$ for evaluation and the \emph{controller network} $Q$ for selection of actions. Furthermore, instead of using a~single output layer for all actions, BDQ introduces multiple branches, each responsible for a~specific subset of actions. 
%For example, in a~robot manipulation task, one branch might control one arm movements, another the other one, another may control the grip etc.
This decomposition allows for parallel processing and shared representations, making it more efficient for the agent to learn and execute tasks with a~diverse action space. BDQs avoid overestimating Q-values, can more rapidly identify action redundancies, and generalise more efficiently by learning a~general Q-value that is shared across many similar actions.

The BDQ loss function is the mean squared temporal-difference error across the branches. Formally, the temporal-difference target is defined as:
\begin{align*}
Y_t^{\textrm{BDQ(d)}} \triangleq R_{t+1} + \gamma Q_d^{-}(S_{t+1}, \argmax_{a \in \mathcal{AC}} Q_d(S_{t+1}, a; \theta_t);\theta_t^{-}), 
\end{align*}
where $Q_d^{-}$ and $Q_d$ denote the branch $d$ of the target and controller network, respectively. Then the loss function is given by:
\begin{align*}
\mathcal{L}_t^{\textrm{BDQ}} \triangleq \mathbb{E}_{(s,a,r,s')\sim \mathcal{D}} \left[ \frac{1}{N}\sum_d (Y_t^{\textrm{BDQ(d)}} - Q_d(s,a_d; \theta_t))^2 \right],
\end{align*}
where $a$ denotes the joint-action tuple $(a_1,a_2,\ldots, a_n)$ and $\mathcal{D}$ is a~(prioritised) experience replay buffer.

\subsection{Graph Neural Networks}
Graph Neural Networks (GNNs) have gained significant attention in recent years due to their ability to model and analyse graph-structured data. Graphs are powerful generic mathematical objects that capture relationships and dependencies between entities in various domains, including social networks, biological systems, and recommendation engines. The main idea underlying GNNs is the use of pairwise message passing, i.e. graph nodes iteratively update their representations by exchanging information with their direct neighbours. In this study, we focus on a~specific type of GNNs, i.e. Graph Convolutional Networks (GCNs). GCNs are a~subclass of GNNs and can be considered a~generalization of classical convolutional neural networks (CNN) to graph-structured data. The essence of GCNs lies in their ability to perform graph convolution, a~key operation that allows them to learn and propagate information across the nodes of a~graph. Graph convolution is inspired by traditional convolutional operations in image processing. While convolutional layers in traditional neural networks operate on grid-structured data, graph convolutional layers extend this concept to non-grid structures. The goal is to capture and aggregate information from neighbouring nodes, enabling the model to understand the relational structure within the graph. The information propagation happens through kernels --~that is functions similar to kernels in CNNs, but adapted to the graph's structure.

\section{Control problem formulation}
\label{sec:control_problem}

We proceed to formulate a~novel target-control problem tailored for cellular reprogramming, understood as an~artificial change of the identity or functional state of a~cell by manipulating its gene expression profile. 

Often biologists are interested not in reaching a~single, fully specified state, but rather a~set of states in which a~specified subset of genes attains a~certain configuration of expression levels, which remains constant afterward. For example, they may be interested in achieving any state in which apoptosis occurs. Therefore, we assume that the target is given  as a~combination of desired values for a~subset of genes rather than a~fully specified attractor as in~\cite{ours_tcs}. We define an~\emph{attractor-target control strategy} as follows.

\begin{figure*}[ht]
    \centering
    \begin{tikzpicture}
        % Define nodes
        \node[circle, fill=blue, inner sep=2pt, label=south west:$\mathcal{A}_S$] (source) at (-9,1.2) {};
        \node[circle, fill=green, inner sep=2pt] (intermediate0) at (-6,1.2) {};
        \node[circle, fill=green, inner sep=2pt] (intermediate) at (-2.1,1) {};
        \node[circle, fill=red, inner sep=2pt, label=below:$\mathcal{A}_{T_1}$] (target) at (2, 1.3) {};
        \node[circle, fill=red, inner sep=2pt, label=right:$\mathcal{A}_{T_2}$] (target2) at (-0.6, 1.7) {};
        \node[circle, fill=red, inner sep=2pt, label=south east:$\mathcal{A}_{T_3}$] (target3) at (3.6, 0.8) {};
        \node[circle, fill=gray, inner sep=2pt, opacity=0.6] (basin0) at (-7.5,1.5) {};
        \node[circle, fill=gray, inner sep=2pt, opacity=0.6] (basin1) at (-3,1) {};
        \node[circle, fill=gray, inner sep=2pt, opacity=0.6] (basin2) at (0,0.5) {};
        \node[circle, fill=gray, inner sep=2pt, opacity=0.6] (basin0_1) at (-7.1,1.2) {};
        \node[circle, fill=gray, inner sep=2pt, opacity=0.6] (basin0_2) at (-6.5,1.8) {};
        \node[circle, fill=gray, inner sep=2pt, opacity=0.6] (basin1_1) at (-2.7,1.4) {};
        \node[circle, fill=gray, inner sep=2pt, opacity=0.6] (basin2_1) at (1.4,0.4) {};

        \node[circle, fill=gray, inner sep=2pt, opacity=0.6] (outer1) at (-8,2) {};
        \node[circle, fill=gray, inner sep=2pt, opacity=0.6] (outer2) at (-5.5,1) {};
        \node[circle, fill=gray, inner sep=2pt, opacity=0.6] (outer3) at (-5,2.2) {};
        \node[circle, fill=gray, inner sep=2pt, opacity=0.6] (outer4) at (-3,2) {};
        \node[circle, fill=gray, inner sep=2pt, opacity=0.6] (outer5) at (-1,0.5) {};

        % Draw interventions
        \node[] at (-8.8,1.7) {\textcolor{red}{\Huge{\Lightning}}};
        \node[] at (-5.8,1.7) {\textcolor{red}{\Huge{\Lightning}}};
        \node[] at (-1.9,1.5) {\textcolor{red}{\Huge{\Lightning}}};
        
        % Draw control paths
        \draw[->, thick, red] (source) to[out=-20,in=220] node[midway, below, text=black, font=\small] {control} (basin0);
        \draw[->, thick, red] (intermediate0) to[out=20,in=160] node[midway, above, text=black, font=\small] {control}(basin1);
        \draw[->, thick, red] (intermediate) to[out=0,in=160] node[midway, above, text=black, font=\small] {control}(basin2);
        \draw[->, dashed, black] (basin0) to[out=40,in=200] (intermediate0);        
        \draw[->, dashed, black] (basin1) -- (intermediate);
        \draw[->, dashed, black] (basin2) to[out=-10,in=160] (target);

        % Enclosing basin ellipses
        \draw[thick] (1.1, .8) ellipse (1.5 and .8) node[below, xshift=1cm, yshift=-0.9cm, text=black, font=\small] {$\mathcal{A}_{T_1}$ basin};
        \draw[thick] (-6.8,1.4) ellipse (1.1 and 0.7) node[below, yshift=-0.65cm, text=black, font=\small] {intermediate basin};
        \draw[thick] (-2.7,1.1) ellipse (1 and 0.6) node[below, yshift=-0.65cm, text=black, font=\small] {intermediate basin};

        \draw[fill=OliveGreen] (-9.1, 1.4) rectangle (-9.3, 1.6);
        \draw[fill=Aquamarine] (-9.1, 1.6) rectangle (-9.3, 1.8);
        \draw[fill=Aquamarine] (-9.1, 1.8) rectangle (-9.3, 2.0);
        \draw[fill=Aquamarine] (-9.1, 2.0) rectangle (-9.3, 2.2);

        \draw[fill=OliveGreen] (-6.1, 1.4) rectangle (-6.3, 1.6);
        \draw[fill=OliveGreen] (-6.1, 1.6) rectangle (-6.3, 1.8);
        \draw[fill=OliveGreen] (-6.1, 1.8) rectangle (-6.3, 2.0);
        \draw[fill=Aquamarine] (-6.1, 2.0) rectangle (-6.3, 2.2);

        \draw[fill=Aquamarine] (-2.2, 1.2) rectangle (-2.4, 1.4);
        \draw[fill=Aquamarine] (-2.2, 1.4) rectangle (-2.4, 1.6);
        \draw[fill=Aquamarine] (-2.2, 1.6) rectangle (-2.4, 1.8);
        \draw[fill=Aquamarine] (-2.2, 1.8) rectangle (-2.4, 2.0);
        
        \draw[fill=Aquamarine] (2.1, 1.5) rectangle (2.3, 1.7);
        \draw[fill=Aquamarine] (2.1, 1.7) rectangle (2.3, 1.9);
        \draw[fill=OliveGreen] (2.1, 1.9) rectangle (2.3, 2.2);
        \draw[fill=Aquamarine] (2.1, 2.1) rectangle (2.3, 2.3);

        \draw[fill=Aquamarine] (-0.5, 1.9) rectangle (-0.3, 2.1);
        \draw[fill=Aquamarine] (-0.5, 2.1) rectangle (-0.3, 2.3);
        \draw[fill=OliveGreen] (-0.5, 2.3) rectangle (-0.3, 2.5);
        \draw[fill=OliveGreen] (-0.5, 2.5) rectangle (-0.3, 2.7);

        \draw[fill=OliveGreen] (3.7, 1.0) rectangle (3.9, 1.2);
        \draw[fill=Aquamarine] (3.7, 1.2) rectangle (3.9, 1.4);
        \draw[fill=OliveGreen] (3.7, 1.4) rectangle (3.9, 1.6);
        \draw[fill=Aquamarine] (3.7, 1.6) rectangle (3.9, 1.8);
        % \node[square, fill=green, inner sep=2pt, label={ }] (target_bar1) at (6.1,1.4) {};
        % \pic {lightning={0,2}{0,0}};

        % Draw target configuration
        \draw[fill=white] (-11.4,-0.3) rectangle (-11.2,-0.1);
        \draw[fill=Aquamarine] (-11.4,-0.1) rectangle (-11.2,0.1);
        \draw[fill=OliveGreen]   (-11.4,0.1) rectangle (-11.2,0.3);
        \draw[fill=white] (-11.4,0.3) rectangle (-11.2,0.5);
        \node[text width=5cm] at (-8.5,-0.2) {Target configuration};

    \end{tikzpicture}
    \caption{Schematic illustration of the concept behind Attractor-Target Control. The source attractor $\mathcal{A}_S$ (blue dot) of a~4-gene network is not aligned with the target configuration specifying the target expression of two genes (expression profile encoding: olivegreen -~gene OFF, aquamarine -~gene ON, white -~gene expression profile unspecified, i.e. either OFF or ON). The target can be any of the attractors (red dots) containing a~state that is aligned with the target configuration. Transitions that follow the original network dynamics are shown with dashed arrows. Red lightnings indicate interventions (control actions) and red arrows indicate the change of state due to interventions. The depicted control strategy is of length 3, which from $\mathcal{A}_S$ traverses two intermediate attractor states (green dots) to finally reach one of the target attractors. Ellipses represent basins of attraction of the respective attractors on the control trajectory (path).}
    \label{fig:atcs}
\end{figure*}

\begin{definition}[Attractor-Target Control Strategy]
\label{def:control_strategy}
Given a~BN, a~source (pseudo-)attractor state, and a~specified target configuration for a~subset of genes, an~\emph{attractor-target control strategy} is a~sequence of instantaneous interventions (perturbations) which guides the BN dynamics from the source (pseudo-)attractor state to some (pseudo-)attractor state that aligns with the desired gene configuration. An~instantaneous intervention is understood as a~simultaneous flip of values for a~subset of genes in a~particular network state that is being perturbed. This results in repositioning the dynamics to a~new state in the state space of the BN/PBN. Instantaneous means that directly after the relocation in the state space, the BN/PBN evolves in accordance with its original dynamics starting from the new state. The application of interventions is limited to (pseudo-)attractor states.

Furthermore, the \emph{length of a~control strategy} is defined as the number of interventions in the control sequence. We refer to a~target control strategy of the shortest length as the \emph{minimal attractor-target control strategy}.
\end{definition}

The concept of the attractor-target control strategy is schematically illustrated in Figure~\ref{fig:atcs} and an~example is provided in Appendix~\ref{app:bn_example}. Notice that, according to the above definition, we aim to guide the network dynamics toward any state within a~subset of (pseudo-)attractor states that match the target configuration, rather than a~specific (pseudo-)attractor. This definition is more general than the definition of the `Attractor-Based Control Strategy' introduced in~\cite{ours_tcs}, where the system is expected to reach a~specific attractor. The current definition of the `Attractor-Target Control Strategy' aligns better with contemporary biological practice.

We now define the \emph{attractor-target control problem}, which is a~variation of the `target-based sequential instantaneous control (ASI) problem defined for BNs in~\cite{SP20a}. An~exact `divide-and-conquer'-type algorithm for solving the ASI problem for BNs is implemented in the software tool CABEAN~\cite{SP21}, which we use as a~benchmark in the evaluation of our proposed approach.

\begin{definition}[Attractor-Target Control]
Given a~BN, a~source (pseudo-)attractor, and a~target configuration for a~subset of genes, find a~minimal attractor-target control strategy.
\end{definition}

\section{GATTACA: Graph-based Attractor-Target Control Algorithm}
\label{sec:algorithm}

We propose a~Graph-based Attractor-Target Control Algorithm (GATTACA) computational framework --~a~GNN-based DRL approach for solving the attractor-target control problem. With cellular reprogramming in mind, we focus on cellular phenotypic functional states, which are observable in experimental practice, and thus allow the DRL agent, which we refer to as the Attractor-Target Graph-based Controller, to take actions only in (pseudo-)attractor states (PA states for short) in agreement with our control problem formulation in Section~\ref{sec:control_problem}. Since the identification of all attractor states is an~NP-hard problem in itself, we improve the pseudo-attractor states identification procedure (PASIP), which was previously established in~\cite{ours_tcs}. The improved PASIP and the Attractor-Target Graph-based Controller constitute the two main components of the GATTACA framework. The source code of our framework together with the trained models are available at~\cite{GATTACA_git_am}.

\subsection{Improved PASIP (iPASIP)}
\label{sec:pasip}
We observed that the original PASIP method in~\cite{ours_tcs} can be fooled by the presence of large attractors in a~BN model. For example, the T-LGL network of~\cite{tlgl} has an~attractor with more than 1000 states. Of course, if the existence of a~large attractor is foreseen and a~rough estimate of its size is available, the PASIP $k_1$ and $k_2$ hyperparameters could be adjusted appropriately based on Theorem~\ref{th:p-a_size_bounds} presented in Appendix~\ref{app:iPASIP_hyperparams}, which was originally proved in~\cite{ours_tcs}. However, this would imply the necessity of longer simulations of the BN model, which are computationally expensive. If the existence of a~large attractor is not anticipated and the hyper-parameters are not adjusted, the original PASIP is prone to getting trapped in such an~attractor while waiting for one of the attractor states to demonstrate a~pronounced frequency of re-visits. To handle such cases effectively, we modify PASIP by adding additional checkpoints that enable the identification of a~representative state within a~large attractor. The improved PASIP (iPASIP) consists of steps executed in two phases: (I) environment pre-processing and (II) DRL agent training. It extends the original PASIP from~\cite{ours_tcs} by incorporating Steps I-2 and II-3 as additional checkpoints.

\begin{enumerate}
    \item[I-1] During environment pre-processing, $k$ trajectories are simulated, each starting from a~randomly selected initial state. Each BN simulation is run for initial $n_0=200$ time steps, which are discarded, i.e. the so-called \emph{burn-in period}. Then, the simulation continues for $n_1=1,000$ time steps during which the visits to individual states are counted. All states in which at least $k_1=5\%$ of the simulation time $n_1$ is spent are added to the list of PA states.
    
    \item[I-2] If at any point more than $d_1=1,000,000$ steps are taken without encountering a~dominating state (one revisited at least $k_1=5\%$ of the time), it is assumed that the dynamics have become trapped in a~large attractor. The most frequently encountered state is then added to the list of PA states as a~representative of this plausible attractor.

    \item[II-1] The simulation of the BN environment may enter a~fix-point attractor not detected in Step~I. If the simulation gets stuck in a~particular state for $n_2=1000$ steps, the state is \AM{verified to be a~fixed-point attractor} and added to the list of PA states.
    
    \item[II-2] During training, the simulation of the BN environment may enter a~multi-state attractor that was not detected in Step~I. For this reason, a~history of the most recently visited states is maintained. When the history buffer reaches the size of $n_3=10,000$ items, revisits for each state are counted and states revisited more than $k_2=15\%$ of times are added to the list of PA states. If no such state exists, the history buffer is cleared and the procedure continues for another $n_3$ time steps. The new PA states are added provided no known PA state was reached. Otherwise, the history information is discarded.

    \item[II-3] As in I-2, if at any point more than $d_2=1,000,000$ steps are taken without encountering a~dominating state (one revisited at least $k_1=5\%$ of the time), it is assumed that the dynamics have become trapped in a~large attractor and a~representative of this plausible attractor is added to the list of PA states.

\end{enumerate}

iPASIP features a~number of hyperparameters, i.e. $n_0$, $n_1$, $n_2$, $n_3$, $d_1$, $d_2$, $k_1$, and $k_2$. Their values should be set on a~case-by-case basis in accordance with prior knowledge about the system under study, or some preliminary experiments, and Theorem~\ref{th:p-a_size_bounds} in Appendix~\ref{app:iPASIP_hyperparams}. The values provided above were suitable for our evaluations discussed in Section~\ref{sec:experiments}.

We notice that while the new checkpoints may clearly introduce false-positive PA states, this occurs rarely. Importantly, the representative states of plausible large attractors stand out due to being frequently revisited. Even if these states are not exact PA states, to a~large extent they resemble them. Therefore, given their rarity, adding them to the list of PA states is acceptable also from the biological point of view. This is confirmed by the results of our case studies, presented and discussed in Section~\ref{sec:results}. 

Since iPASIP can introduce spurious PA states, we shall henceforth use the term `PA state' in a~broader sense: it will refer to all states returned by iPASIP rather than only those defined in Definition~\ref{def:pseudo-attractor}.

\subsection{Attractor-Target Graph-based Controller (ATGC)}

\begin{figure*}[htp]
  \centering
  \includegraphics[width=.8\textwidth]{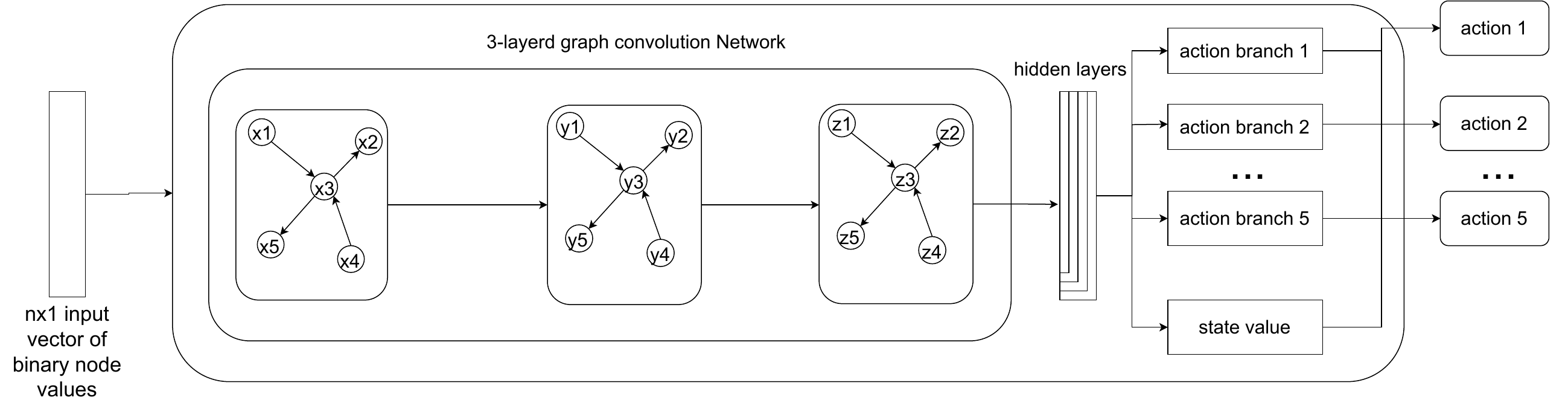}
  \caption{Schematic representation of the ATGC DRL agent of the GATTACA framework.}
  \label{fig:arch}
\end{figure*}

Our DRL agent, which we refer to as the Attractor-Target Graph-based Controller (ATGC), consists of three main components: a~GCN, a~multi-layer perceptron (MLP), and a~BDQ network. The architecture of ATGC is schematically illustrated in Figure~\ref{fig:arch}.

The GCN consists of three graph convolutional layers. In our implementation, a~graph convolutional layer takes the values of each input node from the previous layer and produces the corresponding output vectors for each node. The convolutional layer implements the function:
$$ x'_v = \sum_{u \in N(v)} h_\Theta (v || u - v), $$
where $v$ is the current node, $N(v)$ is the set of neighbours of $v$, and $h$ is the graph convolutional kernel with learnable parameters $\Theta$. In our case, $h$ consists of a~neural network with a~64-neuron deep layer and ReLu activations. Here $h(x || y)$ means conditioning-by-intervention, that is the distribution of $x$ with $y$ set to some fixed value. Its role is to reduce spurious correlations present in the graph; for more details, we refer to~\cite{Wang2019}.

The reason for introducing the GCN into the architecture of the ATGC is twofold. First, to allow the DRL agent to effectively learn the interactions between the BN nodes that represent the system components. In particular, the available knowledge of the structure of the BN is explicitly incorporated into the ATGC by defining the GCN architecture in accordance with the structure graph of the BN. Then, during training, the BN is simulated in accordance with its Boolean functions. Since the resulting state of the BN is observed and used to compute the reward function discussed in the following, the information on the BN dynamics is implicitly encoded via training into the graph convolutional kernels $h_\Theta$ and the remaining parameters of the ATGC architecture.

Second, the GCN creates an~embedding of a~discrete graph into a~continuous space, which is more suitable for machine learning algorithms and eases the training of the DRL agent.
%We expect that our network will learn interactions between genes, and create an embedding of a discreet graph into continuos space

The output of the graph convolutions is passed through a~three-hidden-layers MLP with 1024, 512, and 256 neurons per layer. 

Finally, to allow simultaneous perturbation of a~combination of genes within a~DRL action as in~\cite{ours_tcs}, the MLP is followed by a~BDQ network architecture~\cite{bdq}\footnote{An~architecture consisting of the MLP followed by a~DDQN was originally proposed in~\cite{MCSW22}. In~\cite{ours_tcs}, we replaced the DDQN with the BDQ architecture to allow multiple perturbations of genes within a~single DRL action. For further details, refer to~\cite{ours_tcs}.}.
%The second and third component were in \cite{MCSW22}, but we replace the DDQN architecture of~\cite{MCSW22} with the BDQ~\cite{bdq} in \cite{ours_tcs} and here.
Each of the five action branches and the state value layer in Figure~\ref{fig:arch} is a~three-layer MLP with 256, 512, and 512 neurons. The idea of including the state value layer is to create the internal representation of the current state, so that it can later be included in the calculation of the Q function of the state-action pair. We want all of the branches to share this common representation, as the value of each action changes depending on the state. We consider five action branches as this is the maximum number of nodes in our study that can be perturbed simultaneously within a~single DRL action.

Training in the GATTACA framework is performed as follows. We specify a~source pseudo-attractor state and a~target configuration that determines the desired values for a~subset of genes. We then let ATGC perform actions, i.e. interventions consisting of simultaneous perturbation of up to five genes in a~PA state that is not aligned with the target configuration. After an~action, we let the environment evolve in accordance with its original dynamics into the next PA state. This models the biological process of transforming a~cell from one phenotypic functional state into another by conducting a~series of \emph{in vivo} reprogramming interventions.

We define the DRL reward function as follows:
$$R_a(s, s') = 21 + 100 * \mathbbm{1}_{TA}(s') - |a|,$$
where $s$ and $s'$ are the current and next PA states, respectively, $\mathbbm{1}_{TA}$ is an~indicator function of the target configuration, and $|a|$ is the number of genes perturbed by applying action $a$. 

This reward scheme is balanced to punish the model for perturbing many genes in a~single action while providing enough reward to allow to still reach the target configuration by the means of a~sequence of actions. Our previous experiments showed that it is important that the reward function maintains a~consistent sign during training (in this particular case, it is designed to remain non-negative, which is ensured by the term of 21). \AM{For the loss function, we use the $\mathcal{L}_t^{\textrm{BDQ}}$ function defined in Section~\ref{sec:bdq}}.
%It has also proven useful in \cite{ours_tcs}.

% For the loss function, we use the Double Deep Q Network (DDQN) loss function as in~\cite{MCSW22} and \cite{ours_tcs}, that is the Mean Squared Error (MSE) between the predicted Q-values and the target Q-values, calculated as follows:
% $$
% \mathcal{L}(\theta) = \mathbb{E} \left[ \left( q_{\pi}(s, a) - target(s, a) \right)^2 \right],
% \label{eq:loss}
% $$
% where $target$ is the delayed copy of the $q_\pi$ network.

Our DRL agent is trained in episodes. An~episode is a~sequence of states, actions, and rewards that starts in the specified source state and ends in any of the PA states that comply with the target configuration. It represents a~complete trajectory (path) of the agent's interaction with the environment. 

For each episode, we randomly choose a~source PA state. We make sure that this state does not satisfy the target configuration. Then we let our agent apply its strategy. If applying 100 actions within an~episode is not sufficient to reach the target configuration, the respective episode is considered unsuccessful and terminated.

\section{Experiments}
\label{sec:experiments}

We evaluate how our GATTACA framework scales with the size of the BN model. We focus on large, verified models taken from the literature for well-known real-world biological networks.

\subsection{Case Studies: Models and Their Target Configurations}
\label{sec:models}

To evaluate the performance of our approach, we select a~variety of models from the literature, ranging in size from 35 nodes (the Bladder model) to 188 nodes (the CD4$^+$ model). To ensure that the target configurations used in our evaluation are both realistic and of significance to biologists, we focus on the configurations considered in the original publications where these models were introduced. The models and their corresponding target configurations are listed in Table~\ref{tab:models}. Below, we briefly present each of the models.
%(Reference column in Tab.~\ref{tab:models}.)

\begin{table}[ht]
\centering
\begin{tabular}{|c|c|c|c|}
\hline
\textbf{Model} & \textbf{BD Id} & \textbf{Size} & \textbf{Target Configuration}\\
\hline
Bladder & \href{https://github.com/sybila/biodivine-boolean-models/tree/main/models/%5Bid-183%5D__%5Bvar-31%5D__%5Bin-4%5D__%5BALTERATIONS-IN-BLADDER%5D}{183} & 35 & Apoptosis\_b1=1 or Apoptosis\_b2=1\\
\hline
MAPK & \href{https://github.com/sybila/biodivine-boolean-models/tree/main/models/%5Bid-070%5D__%5Bvar-49%5D__%5Bin-4%5D__%5BMAPK-CANCER-CELL-FATE%5D}{70} & 53  & Apoptosis=1\\ 
\hline
T-LGL & \href{https://github.com/sybila/biodivine-boolean-models/tree/main/models/%5Bid-014%5D__%5Bvar-54%5D__%5Bin-7%5D__%5BT-LGL-SURVIVAL-NETWORK-2008%5D}{14} & 61 & Apoptosis=1\\ 
\hline
Bortezomib & \href{https://github.com/sybila/biodivine-boolean-models/tree/main/models/%5Bid-059%5D__%5Bvar-62%5D__%5Bin-5%5D__%5BBORTEZOMIB-RESPONSES-IN-MYELOMA-CELLS%5D}{59} & 67 & JNK=0, p21=Cas3=1\\ 
\hline
T-diff & - & 68 / 71 & TBET=RORGT=FOXP3=1, GATA3=0\\ 
\hline
ABA & \href{https://github.com/sybila/biodivine-boolean-models/tree/main/models/%5Bid-084%5D__%5Bvar-58%5D__%5Bin-23%5D__%5BABA-INDUCED-STOMATAL-CLOSURE%5D}{84} & 81 & Closure=1\\ 
\hline
MCF-7 & \href{https://github.com/sybila/biodivine-boolean-models/tree/main/models/%5Bid-232%5D__%5Bvar-109%5D__%5Bin-8%5D__%5BBREAST-CANCER-PATHWAYS-MCF7-COMPLETE%5D}{232} & 117 &  CASP3=E2F1=1\\ 
\hline
CD4$^+$ & \href{https://github.com/sybila/biodivine-boolean-models/tree/main/models/%5Bid-050%5D__%5Bvar-154%5D__%5Bin-34%5D__%5BCD4-T-CELL-SIGNALING%5D}{50} & 188 & GATA3=1\\ 
\hline
%Random200 & 200 & x3=1 \\
%\hline
\end{tabular}
\caption{BN models of real-world biological networks with their target configurations used for the performance evaluation of the GATTACA framework. The BD Id column provides model identification numbers in the BioDivine repository~\cite{BioDivine}.}
\label{tab:models}
\end{table}

\begin{table*}[ht]
\centering
    \begin{tabular}{|c|c|>{\raggedright\arraybackslash}p{10.2cm}|}
\hline
\textbf{Model} & \textbf{Size (Inputs)} & \textbf{Environmental Condition}\\ 
\hline
Bladder & 35 (4) & EGFR\_stimulus = 0, DNAdamage = *, FGFR3\_stimulus = *, GrowthInhibitors = 1\\
\hline
MAPK & 53 (4) & EGFR\_stimulus = 0, FGFR3\_stimulus = 0, Growth\_Arrest = *, DNA\_damage = *\\
\hline
T-LGL & 61 (7) & CD45 = *, IFN = *, IL15 = *, PDGF = *, Stimuli = *, Stimuli2 = *, TAX = *\\
\hline
Bortezomib & 67 (5) & XX = 0, SHP1 = 0, TNFAR = 0, Bort = 0, TNFA = 0\\
\hline
T-diff & 68 / 71 (24) & APC = 1, IFNB\_e = 0, IFNG\_e = 0, IL2\_e = 1, IL4\_e = 0, IL6\_e = 0, IL10\_e = 0, IL12\_e = 0, IL15\_e = 0, IL21\_e = 0, IL23\_e = 0, IL27\_e = 0, TGFB\_e = 1, IFNGR1 = 1, IFNGR2 = 1, GP130 = 1, IL6RA = 1, CGC = 1, IL12RB2 = 1, IL10RB = 1, IL10RA = 1, IL15RA = 1, IL2RB = 1, IL27RA = 1\\
\hline
ABA & 81 (23) & PC = *, PtdInsP3 = *, DAGK = *, CPK23 = *, NAD = *, Sph = *, SCAB1 = *, Nitrite = *, ERA1 = *, ABH1 = *, SPP1 = *, GEF1\_4\_10 = *, ARP\_Complex = *, MRP5 = *, RCN1 = *, NtSyp121 = *, ABA = *, NADPH = *, GCR1 = *, PtdInsP4 = *, GAPC1\_2 = *, GTP = *, CPK6 = *\\
\hline
MCF-7 & 117 (8) & DLL\_i = 1, EGF = 1, ES = 1, IGF1 = 1, INS = 1, NRG1 = 1, PG = 1, WNT1 = 1\\
\hline
CD4$^+$ & 188 (34) & IL27RA = 0, IL27\_e = 0, GP130 = 0, Galpha\_QL = 0, IL2RB = 0, CGC = 0, Galpha\_iL = 0, MHC\_II = 0, APC = 0, IL18\_e = 0, IL9\_e = 0, IFNB\_e = 0, ECM = 0, IL21\_e = 0, alpha\_13L = 0, IL10RA = 0, IL10RB = 0, IL10\_e = 0, IL15\_e = 0, B7 = 0, IFNGR1 = 0, IFNGR2 = 0, IFNG\_e = 0, CAV1\_ACTIVATOR = *, GalphaS\_L = 0, IL4\_e = 0, IL6\_e = 0, IL6RA = 0, TGFB\_e = 0, IL22\_e = 0, IL2\_e = 0, IL23\_e = 0, IL15RA = 0, IL12\_e = 0\\
\hline
%Random200 & 200 & x3=1 \\
%\hline
\end{tabular}
\caption{Environmental conditions for the BN models of real-world biological networks used in the evaluation of GATTACA. Unspecified node values are indicated with *.}
\label{tab:models_env_conds}
\end{table*}

\paragraph{Bladder (Bladder)}
The logical model for the prediction of bladder cancer cells to become invasive was originally introduced in~\cite{bladder} with the purpose of understanding how genetic alterations combine to promote cancer tumorigenesis. Specifically, the focus is on the relationships between the alterations, i.e. mutual exclusivity and co-occurrence. Using statistical analysis on tumour datasets and a~logical modelling approach, the study explores patterns of genetic alterations that affect growth factor receptor signalling, cell cycle regulation, and apoptosis. The model confirms conditions where genetic alterations drive tumour progression and suggests that additional mutations, such as PIK3CA or p21CIP deletion, may enhance invasiveness.

\paragraph{MAPK cancer cell fate (MAPK)}
The Mitogen-Activated Protein Kinase (MAPK) network is a~system of tightly interconnected signalling pathways involved in diverse cellular processes, such as cell cycle, differentiation, survival, and apoptosis. In~\cite{mapk}, a~comprehensive and generic reaction map of the MAPK signalling network is constructed based on an~extensive analysis of published data. To investigate MAPK responses to different stimuli and understand their involvement in cell fate decisions in urinary bladder cancer, specifically the balance between cell proliferation and cell death, a~logical model comprising 53 nodes is constructed, incorporating the most crucial components and interactions. The Boolean model is verified to be globally consistent with published data and it is capable of recapitulating salient dynamical properties of the complex biological network.

\paragraph{T-LGL survival network (T-LGL)}
T~cell large granular lymphocyte (T-LGL) leuke\-mia features a~clonal expansion of antigen-primed, competent, cytotoxic T~lymphocytes (CTL). To understand the signalling components that are crucial for the survival of CTL in T-LGL leukemia, a~T-LGL survival signalling network is constructed in~\cite{tlgl}. The network integrates the signalling pathways involved in normal CTL activation and the known deregulations of survival signalling in leukemic T-LGL. The network is translated into a~BN model that  describes the signalling involved in maintaining the long-term survival of competent CTL in humans. The BN model is capable of reproducing a~clinically relevant complex process. In particular, it predicts that the persistence of IL-15 and PDGF is sufficient to reproduce all known deregulations in leukemic T-LGL. For further details, we refer to~\cite{tlgl}.

\paragraph{ABA-induced stomatal closure (ABA)} A~vital process for drought tolerance in plants is examined in~\cite{aba}. To reduce water loss during drought stress conditions, the phytohormone abscisic acid (ABA) triggers the closure of stomata, i.e. microscopic pores on leaf surfaces through which water loss and carbon dioxide uptake occur. Based on an~extensive review of the literature on ABA signal transduction responses underlying stomatal closure, a~network of 84 nodes and 156 edges is constructed. This network is then translated into a~BN model which predictions agree with 82 out of 99 considered experiments. The model reveals nodes that could be controlled to influence stomatal closure and provides a~substantial number of novel predictions that could be subject to wet-lab validation.

\paragraph{MCF-7 (MCF-7)} Cell growth and proliferation in breast cancer incorporating common signalling pathways is analysed in~\cite{mcf7}. To account for the fact that in different cell lines the same combination of drugs can produce different synergistic effects leading to heterogenous responses, the generic model is tailored to represent five breast cancer cell lines. In particular, by integrating information on cell line specific mutations, gene expression profiles, and observed behaviours from drug-perturbed experiments, a~BN model for the MCF-7 cell line is proposed. The model predictions on cell-line specific protein activities and drug-response behaviours are in agreement with experimental data.

\paragraph{T-diff (T-diff)}
A comprehensive 71-node Boolean network model was developed to represent the regulatory and signaling pathways controlling CD4\textsuperscript{+} T helper (Th) cell differentiation \cite{tdiff}. The model captures key transcription factors, cytokine receptors, and intracellular signaling components involved in Th lineage specification. Simulations revealed the emergence of stable states corresponding to canonical Th1, Th2, Th17, and Treg subsets, as well as hybrid cell types that co-express lineage-specific markers, reflecting the context-dependent plasticity of Th cells. Environmental inputs, such as cytokine concentrations, were shown to modulate the attainable cell fates, emphasizing the model’s ability to integrate external cues. A~reduced 68-node version of the original network was subsequently used as a~case study in~\cite{SP20a}, focusing on specific environmental conditions to investigate dynamic transitions between Th phenotypes. This work demonstrated the model’s applicability to studying immune cell adaptation in diverse microenvironments and highlighted potential regulatory checkpoints for therapeutic intervention.

\paragraph{Bortezomib}
Bortezomib is a~commonly used first-line agent in multiple myeloma (MM) treatment. In~\cite{bortezomib}, a~Boolean network model of 67 nodes was constructed based on the literature that includes major survival and apoptotic signalling pathways in U266 MM cells. By combining \emph{in vitro} experiments and computational modelling with pathway-specific inhibitors, the individual and simultaneous roles of NF$\kappa$B and JAK/STAT3 pathways were evaluated. The BN model confirmed that stress accumulation due to proteasome inhibition is a~major pathway of myeloma cell death despite the activation of phospho-NF$\kappa$B protein expression, which contradicted a~major proposed hypothesis of bortezomib effects.

\paragraph{CD4$^+$} 

Caveolin-1 (CAV1) is a~vital scaffold protein heterogeneously expressed in both healthy and malignant tissue. In~\cite{cd4}, the role of overexpressed CAV1 in T-cell leukemia is studied. In particular, a~188-node BN model of a~CD4$^+$ immune effector T-cell is constructed. The model incorporates various signalling pathways and
corresponding protein-to-protein, protein-phosphorylation, and kinase interactions to mimic cellular dynamics and molecular signalling under healthy and immunocompromised conditions. The model was used to predict and examine \emph{in silico} the heterogeneous effects and mechanisms of CAV1. Specifically, to verify the hypothesis that the role of CAV1 in immune synapse formation contributes to immune regulation during leukemic progression. It was verified that the local interactions built into the BN model were able to accurately mimic complex phenomena observed in the laboratory.

\subsection{Performance Evaluation Methodology}

We evaluated the performance of the GATTACA framework in solving the attractor-target control problem formulated in Section~\ref{sec:control_problem} on asynchronous BN models of real biological systems in a~wide range of sizes using case studies described in Section~\ref{sec:models}.

In logical models, input nodes typically define specific environmental conditions and are kept fixed.
%When changing from one environmental condition to another, these nodes always need to be set accordingly. 
Therefore, when going from one attractor in a~given environmental condition to another attractor in a~different condition, the input nodes always need to be set accordingly. Hence, they are not very interesting from the control problem point of view. That is why for the Bortezomib, T-diff, MCF-7, and CD4$^+$ models we consider a~specific, biologically relevant environmental condition taken from the respective publication in which a~particular model was originally introduced. However, for the smaller Bladder and MAPK models, we fix only a~subset of input nodes, which in both cases leads to four environmental conditions being considered at once. This is due to the fact that in the majority of cases where all input nodes are set to a~specific environmental condition, the models manifest either a~single attractor or no or all attractors aligned with the target configuration. Hence, the control problem has either no solution or trivial ones, where no control action is needed. Therefore, we allow more than one fully specified environmental condition to be considered at once by allowing any value for some of the input nodes. 

For the T-LGL and ABA models, no environmental conditions were specified in the publications in which they were introduced. Therefore, we consider all possible environmental conditions for these two models.

Since we want ATGC of our GATTACA framework to learn the relationship between the target genes and other genes, we let it perturb all nodes (including input nodes) except the ones specified in the target configuration, which we refer to as target genes. Beyond this, we do not impose any other restrictions on genes that can be perturbed: we allow full flexibility in selecting genes for interventions to ensure a~fair comparison with the CABEAN benchmark and to demonstrate the potential of the GATTACA framework to identify optimal or suboptimal strategies in large networks. However, we emphasize that our framework can be straightforwardly modified to incorporate prior biological knowledge by restricting perturbations to a~predefined subset of genes. 
%In our experiments, we specify the target genes and their configurations based on the literature from which the models were sourced. The details are presented in Section~\ref{sec:models}.

Furthermore, the more genes that need to be perturbed in a~single intervention, the more challenging and expensive the wet-lab experimental procedure becomes. As a~result, the \emph{in vitro} implementation of the control strategy predicted by the GATTACA framework may be unfeasible. Hence, to make our approach as cost-effective as possible, we decide to limit the number of nodes that can be perturbed simultaneously in a~single intervention. This number is a~parameter of our framework and we set it to five in our experiments. This setting suffices to compute successful control strategies for all of our case studies. Nevertheless, the parameter value can be adjusted to particular needs as required.

ATGC is trained to minimise the number of control actions required to drive a~BN model from a~randomly chosen source attractor to any attractor in which the given target configuration is realised. Specifically, for a~specified environmental condition, a~training episode begins by randomly selecting one of the PA states identified so far by iPASIP that does not align with the given target configuration. Throughout each episode, we track the number of control actions taken, referred to as the \emph{episode control length}, as ATGC guides the network dynamics toward any PA state aligned with the given target configuration. ATGC can take up to 100 control actions per episode. If it fails to control the BN model within this limit, the episode is considered unsuccessful, indicating that no valid control strategy was found.

To evaluate the performance of the GATTACA framework on a~particular BN, we recover control strategies by simulating the environment from a~randomly selected source PA state that is not aligned with the target configuration. At each intermediate PA state, a~control action predicted by trained ATGC is applied. For each recovered control strategy, we record its length and the information whether the target configuration is reached. Due to the nondeterministic dynamics of the environment introduced by the asynchronous update mode, results may vary between runs. To address this issue, we recover a~control strategy 10 times for each of the randomly selected source PA states. For a~given BN model, we report the average length of successful control strategies. 

We compare the results of the GATTACA framework with those obtained using the state-of-the-art exact attractor-based sequential instantaneous source-target control (ASI) algorithm introduced in~\cite{cmsb19} and implemented in the CABEAN software tool~\cite{SP21}. The exact algorithm follows the divide-and-conquer approach based on SCC-decomposition of BN structure graph and it leverages symbolic techniques based on Binary Decision Diagrams (BDDs) to encode the BN dynamics. Amongst the control algorithms provided by CABEAN, we selected ASI as the best match for the control problem considered in this study.

For a~fair comparison, we set in CABEAN the initial values of the input nodes of a~BN model in accordance with a~particular environmental condition. CABEAN requires at least some of the input nodes to have fixed values. In the case of partially specified environmental conditions of Bladder, MAPK, and CD4$^+$ models it is possible to consider in CABEAN all of the corresponding fully specified environmental conditions at once as is the case of our framework.

However, in the case of all environmental conditions considered, e.g. the T-LGL model, fully specified environmental conditions are enumerated and considered one by one in CABEAN. For each such condition, all attractors are computed with CABEAN and divided into source attractors, i.e. those not aligned with the target configuration, and target attractors, i.e. those aligned with the target configuration. Then, for each source attractor, the shortest path to any of the target attractors is computed with CABEAN. If there is no available target attractor for a~given environmental condition, we check all other environmental conditions for a~target attractor. We first switch the environmental condition, and then (if necessary) use CABEAN to control the network within the new environmental condition (the GATTACA framework can switch environmental conditions automatically by letting it perturb input nodes). We provide an~example of this scenario in Appendix~\ref{app:atc_example}. To account for the ``manual'' environmental condition switch, one is added to the control path length returned by CABEAN. We take the minimum length over all other environmental conditions that contain a~target attractor. Finally, the averaged length of the shortest paths is reported over all environmental conditions and source attractors. 

Formally, the average control path length for a~BN model $M$, calculated from the exact results obtained with CABEAN, is denoted $\mathsf{ACPL}_{\mathrm{CABEAN}}(M)$ and is defined in accordance with the following formula:
\begin{strip}
  \begin{align}
    \label{eq:acpl}
    \mathsf{ACPL}_{\mathrm{CABEAN}}(M)
      \;=\;\mathbb{E}_{e \in E_M}\left[\,
        \mathbb{E}_{s \in \mathcal{S}^{M}_e}\left[
          \begin{cases}
            \displaystyle\min_{t \in \mathcal{T}^{M}_e}\!\mathsf{cpl}(s,t)
              & \text{if }\mathcal{T}^{M}_e \neq \emptyset,\\[1ex]
            \displaystyle\min_{e' \in E_{M}/\{e\}}\Bigl(1 + 
              \min_{t \in \mathcal{T}^{M}_{e'}}\!\mathsf{cpl}(s,t)\Bigr)
              & \text{otherwise.}
          \end{cases}
        \right]
      \right],
  \end{align}%
\end{strip}\par\vspace{-1\baselineskip}
\noindent where $E_M$ is the set of all separate environmental conditions considered for model $M$, $\mathcal{S}^{M}_e$ is the set of source attractors, $\mathcal{T}^{M}_e$ is the set of target attractors for a~given environmental condition $e$ for model $M$, and $\mathsf{cpl}(s,t)$ is the control path length computed by CABEAN for a~specific pair of source ($s$) and target ($t$) attractors. Notice that in the case of a~fully or partially specified environmental condition where all the corresponding fully specified conditions can be handled by CABEAN at once, the set $E_M$ in Equation~\eqref{eq:acpl} is a~singleton. In the case where all environmental conditions are considered, the cardinality of $E_M$ is equal to the number of distinct possible settings of the values for all input nodes.

The remaining parameters of the GATTACA framework are set as follows: Adam lr = $1\mathrm{e}{-4}$, $\gamma = 0.99$, batch size = 128, replay = $10^6$, PER $\alpha$ = 0.6, $\beta = 0.4 \to 1.0$, target update $\tau=0.01$, $\epsilon$-greedy ($1.0 \to 0.05$, $10^6$ steps), and gradient clip = 10.

\section{Results}
\label{sec:results}
We evaluate the GATTACA framework on the BN models presented in Section~\ref{sec:models}. The results obtained demonstrate that our approach is scalable --~it is capable of controlling environments across a~wide range of sizes, particularly BN model characterised by huge state spaces.

\subsection{Evaluation of iPASIP}

Our iPASIP method, although it is a~purely statistical method based on model simulations, provides a~strong alternative to traditional methods for finding BN attractor states. Table~\ref{tab:results-iPASIP} presents the summarised results for each model in which CABEAN was able to compute the attractors for the environmental conditions specified in Table~\ref{tab:models_env_conds}.

To provide a~more comprehensive evaluation of iPASIP, we additionally ran it on the Bladder, T-LGL, and Bortezomib models without specifying the environmental condition --~that is, simultaneously considering all of them. For comparison, we computed the attractors for each model with CABEAN in all individual environmental conditions one by one, i.e. in 16, 128, and 32 conditions, respectively. The numbers of attractors in all environmental conditions are presented in square brackets in Table~\ref{tab:results-iPASIP}. 

In the case of the MAPK model, CABEAN failed in 3 of 16 possible environmental conditions. Running iPASIP under all environmental conditions resulted in a~very large number of identified PA states, which we could not verify due to the lack of ground-truth information. However, running iPASIP under all environmental conditions except the ones in which CABEAN failed resulted in fully correct identification of all attractors, see Table~\ref{tab:results-iPASIP}.

\begin{table*}[ht]
\centering
\begin{tabular}{|c|c||c|c|c||c|c|c|}
    \hline
    \multirow{2}{*}{\textbf{Model}} & 
    \multirow{2}{*}{\textbf{Size (Inputs)}} &
    \multicolumn{3}{c||}{\textbf{iPASIP PA States}} &
    \multicolumn{3}{c|}{\textbf{CABEAN}} \\
    \cline{3-8}
    & & \textbf{F-A} & \textbf{C-A} & \textbf{Spurious}
      & \textbf{F-A} & \textbf{C-A} & \textbf{Failed}\\
    \hline\hline
    Bladder & 35 (4) & 9 [20] & 0 [5] & 0 [0] & 9 [20] & 0 [4] & 1 of 16\\ 
    \hline
    MAPK & 53 (4) & 6 [12] & 0 [3] & 0 [0] & 6 [12] & 0 [3] & 3 of 16\\
    \hline
    T-LGL & 61 (7) & 149 & 182 & 2? & 166 & 130 & 6 of 128\\
    \hline
    Bortezomib & 67 (5) & 5 [83] & 0 [0] & 0 [0] & 5 [83] & 0 [0] & 0 of 32\\
    \hline
    T-diff & 68 (24) & 12 & 0 & 0 & 12 & 0 & 0 of 1\\
    \hline
    MCF-7 & 117 (8) & 4 & 0 & 0 & 4 & 0 & N/A\\
    \hline
    CD4$^+$ & 188 (34) & 12 & 0 & 1 & 12 & 0 & 0 of 1\\
    \hline
\end{tabular}
\caption{Comparison of the numbers of pseudo-attractor (PA) states identified by iPASIP with the numbers of attractors obtained with CABEAN. The numbers of fixpoint attractors (F-A), cyclic or complex attractors (C-A), and spurious PA states are presented for the environmental conditions in Table~\ref{tab:models_env_conds} (plain numbers) and for input nodes left unspecified, i.e. all possible environmental conditions (numbers in square brackets). The numbers of environmental conditions for which CABEAN failed to compute a~result are shown in the `Failed' column. The `?' indicates that our verification of whether the PA states are indeed spurious failed.}
\label{tab:results-iPASIP}
\end{table*}

For the Bladder model, iPASIP found 25 PA states in total in all 16 environmental conditions, while CABEAN returned 24 attractors in 15 environmental conditions and failed in one, namely where the \textsf{EGFR\_stimulus} input node is set to 1 and all the remaining three input nodes are set to 0. The 24 CABEAN-found attractors are represented within the 25 PA states. By computing an~attractor reachable from the extra PA state identified by iPASIP, we found that it is in fact an~attractor state of an~attractor consisting of 184,320 states under the environmental condition that CABEAN was unable to handle.

\begin{figure}[ht]
    \centering
    \includegraphics[width=.75\linewidth]{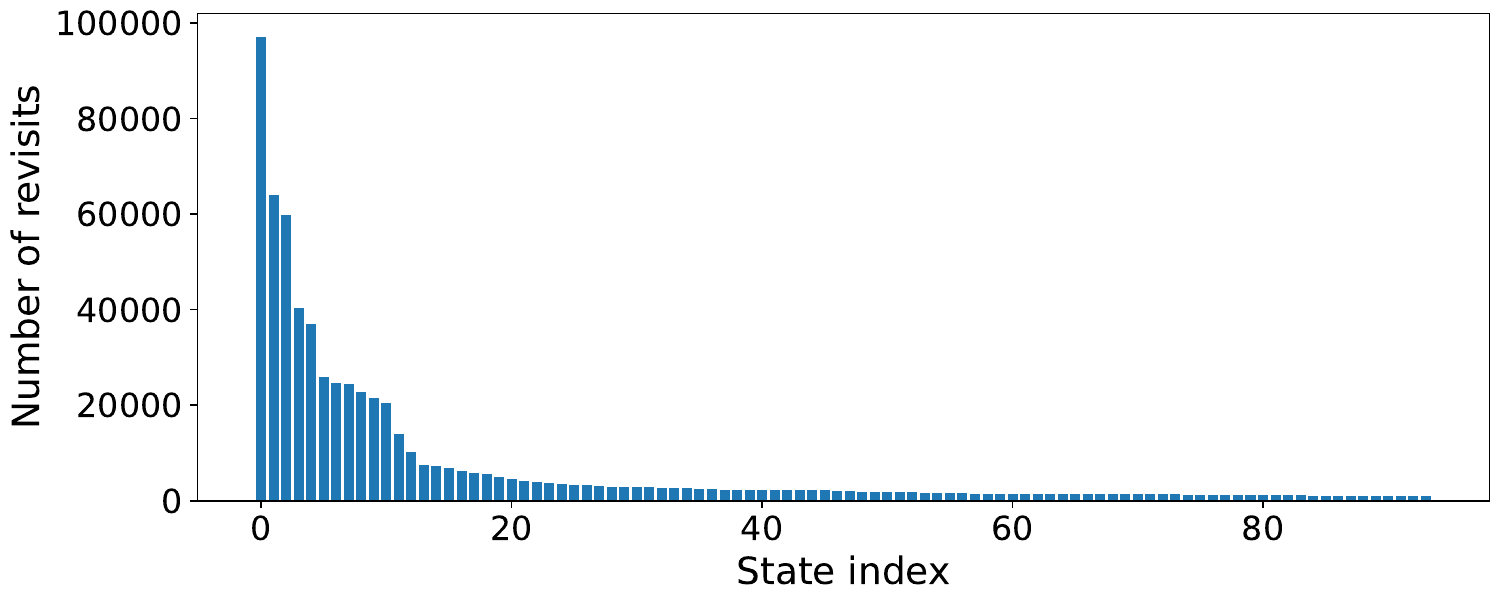}
    \caption{Bar plot of the number of revisits to individual states of the Bladder BN model attractor containing the extra PA state identified by iPASIP under the environmental condition of \textsf{EGFR\_stimulus} set to 1 and all remaining input nodes set to 0. For readability, only states with at least 1,000 revisits are shown in the decreasing order of revisit count. State of index 0 is the PA state found by iPASIP and the remaining attractor states are indexed arbitrarily.}
    \label{fig:bladder_attr_hist}
\end{figure}

We further investigated the distribution of revisits to the individual states of this large attractor. We simulated a~trajectory of 1,000,000 steps with the PA state as the initial state. The numbers of revisits for individual states are shown in Figure~\ref{fig:bladder_attr_hist}. The extra PA state is clearly dominating all other attractor states in terms of the count of revisits and is correctly identified by the iPASIP method despite the attractor being exceptionally large.

In the cases of the T-LGL and CD4$^+$ models, iPASIP overestimates the number of attractor states, i.e. some of the PA states are spurious attractor states, although for the T-LGL model we were unable to confirm that indeed the two states are spurious: we failed to compute attractors reachable from the potentially spurious states with a~graph traversal approach. Nevertheless, as argued in Section~\ref{sec:pasip}, the additional states stand out due to being frequently revisited. Thus, they can still be considered viable for performing control actions. 

The extra PA state\footnote{In the spurious attractor state the nodes \textsf{Cofilin}, \textsf{GATA3}, \textsf{IL4}, \textsf{PI3K}, \textsf{proliferation}, \textsf{AKT}, \textsf{PIP3\_345}, \textsf{Dec2}, \textsf{IRF4}, \textsf{IL4RA}, \textsf{PDK1}, and \textsf{IKB} are 1 and all the remaining nodes are 0.} of the CD4$^+$ BN model is indeed spurious --~this can be inferred from the fact that CABEAN determined all attractor states for the given environmental condition. By simulating 1,000,000-step trajectories from this spurious attractor state, we observed that it was revisited more than 996,000 times and the trajectories consisted of less than 70 distinct states. Clearly, in this sense, the state resembles a~stable state.

We verified that for all models except T-LGL all the attractors found by CABEAN have a~representative amongst the PA states. For T-LGL, 143 of 166 CABEAN fixpoint attractors and 71 out of 130 CABEAN cyclic or complex attractors are detected by iPASIP, some of them multiple times as more than one PA state may belong to the same attractor (data not shown). The remaining fixpoint PA states found by iPASIP are attractor states in the six environmental conditions in which CABEAN failed\footnote{By failure we mean the fact that CABEAN ended with an~error.}.

In summary, the iPASIP approach provides a~scalable and effective solution for identifying true PA states, as defined in Definition~\ref{def:pseudo-attractor}, in large-size BN models. Moreover, as the case studies show, iPASIP often successfully identifies all the attractor states of a~large BN model.

\subsection{Evaluation of the GATTACA framework}

With regard to control problem solutions found by our GATTACA framework, the average length of the control strategies is typically within one control action/intervention of the optimal solution computed in accordance with the formula in Equation~\eqref{eq:acpl} based on the control path lengths obtained by the exact ASI algorithm implemented in the CABEAN software tool, as shown in Table~\ref{tab:results-ATGC}.

However, for the ABA model the number of environmental conditions is too large ($2^{23}$) to be handled by CABEAN (entries marked N/A in Table~\ref{tab:results-ATGC}). As obtained with the GATTACA framework, the ABA model is characterised by a~very large number of PA states. However, the GATTACA framework successfully identifies effective control strategies, demonstrating the scalability of our framework in this respect. Moreover, we conducted experiments in which the number of PA states was overestimated by a~factor of 1000 and ATGC was still able to identify effective control strategies (data not shown). Those two facts prove that the ATGC is resilient to both high attractor counts and spurious PA states.

\begin{table*}[ht]
\centering
\begin{tabular}{|c|c|c|c|c|c|}
    \hline
    \textbf{Model} & \textbf{Size (Inputs)} & \textbf{PA states} & \textbf{Attractors} & \textbf{GATTACA} & \textbf{CABEAN}\\ %[0.5ex]
    \hline\hline
    Bladder & 35 (4) & 9 & 9 & 1.0 & 1.0\\
    \hline
    MAPK & 53 (4) & 6 & 6 &  1.0 & 1.0\\ 
    \hline
    T-LGL & 61 (7) & 333 & 296 & 1.02 & 1.0\\   
    \hline
    Bortezomib & 67 (5) & 5 & 5 & 1.99 & 1.05\\
    \hline
    T-diff & 68 (24) & 12 & 12 & 2.55 & 1.18\\
    \hline
    ABA & 81 (23) & 40,000 & N/A & 2.23 &N/A\\
    \hline
    MCF-7 & 117 (8) & 4 & 4 & 2.33 & N/A\\
    \hline
    CD4$^+$ & 188 (34) & 13 & 12 & 1.16 & 1.03\\
    \hline\hline
    Random200 & 200 (0) & 165 & N/A & 7.1 & N/A\\
    \hline
\end{tabular}
\caption{Comparison of the average control strategy lengths required to drive the network from the source state to the target configuration obtained with the GATTACA framework and the exact ASI method implemented in CABEAN. Results that could not be obtained with CABEAN are indicated as not available (N/A).}
\label{tab:results-ATGC}
\end{table*}

To further investigate the scalability potential of the GATTACA framework, we analysed the MCF-7 model under all possible environmental conditions. As~in the case of the ABA model, MCF-7 is characterised by a~large number of PA states and the GATTACA framework is still capable of successfully identifying control strategies, as shown in Table~\ref{tab:tdiff}. Unfortunately, we were unable to obtain exact results where all $256$ fully specified environmental conditions were considered with CABEAN as it could not complete the attractors calculations in 168 hours (one week) on an~8-Core 2.25 GHz AMD EPYC 7742 processor with 256 GB of RAM.

Next, we analysed the original T-diff model with 71 nodes, introduced in~\cite{tdiff}, under two environmental conditions: 1)~the original, partially specified (PS) condition from~\cite{tdiff} and 2)~the fully specified (FS) condition from~\cite{SP20a} presented in Table~\ref{tab:models_env_conds}. The former condition leaves some of the input nodes unspecified with respect to the latter, i.e. IFNGR1, IFNGR2, GP130, IL6RA, CGC, IL12RB2, IL10RB, IL10RA, IL15RA, IL2RB, and IL27RA.
%We refer to these conditions as the partially and fully specified environmental conditions, respectively.

The original 71-node T-diff model, under both conditions, is characterised by a~large number of PA states identified by iPASIP, i.e. 156,864 and 25,856 states, respectively. CABEAN fails to identify the attractors and cannot compute control strategies for these BNs. However, the GATTACA framework successfully identifies control strategies in both cases. The average control strategy length is 6.17, as shown in Table~\ref{tab:tdiff}. It is noticeably larger than for all other BN models considered in this study. Nevertheless, this is justifiable, as such large numbers of PA states suggest that the two BN models have large numbers of attractors. Meanwhile, the size of the original T-diff model (71 nodes) is comparable to the size of the model in~\cite{SP20a} (68 nodes), the latter with only 12 attractors; see Table~\ref{tab:results-ATGC}. The state space of the original T-diff model under the two environmental conditions is then highly probable to be divided into much finer-grained basins of attraction than in the case of the modified model in~\cite{SP20a}. Consequently, driving the dynamics towards a~PA~state aligned with the given target configuration requires an~action that places the network dynamics into a~relatively small basin of attraction, likely necessitating simultaneous perturbation of many genes. Since this is restricted due to the limit imposed on the maximum number of genes that can be perturbed at once within a~single action, identified control strategies must traverse multiple attractors to reach the target configuration.

\begin{table}[ht]
\centering
\begin{tabular}{|l|c|c|c|}
\hline
\textbf{Model} & \textbf{Size} & \textbf{PA states} & \textbf{GATTACA} \\ 
\hline
MCF-7 & 117 & 1,500 & 2.11\\
\hline
Original T-diff in PS env. cond. & 71 & 156,864 & 6.17\\
\hline
Original T-diff in FS env. cond. & 71 & 25,856 & 6.17\\
\hline
%modified with modified env. condition & 68 & 12 & 2.55 \\
%\hline
\end{tabular}
\caption{Average control strategy lengths obtained with the GATTACA framework for the MCF-7 model under all environmental conditions and the original 71-node T-diff model of~\cite{tdiff} under two environmental conditions (env. cond.), i.e. the partially specified (PS) one in~\cite{tdiff} and the fully specified (FS) one in~\cite{SP20a}. Both conditions are characterised by large number of PA states identified by iPASIP.}
\label{tab:tdiff}
\end{table}

Finally, we performed one additional scalability test of the GATTACA framework on a~randomly generated BN model of 200 nodes (Random200). The model is challenging as it has no input nodes and its structure graph consists of a~single SCC. Therefore, the attractor detection algorithm and the ASI control algorithm in CABEAN cannot exploit the SCC-decomposition and CABEAN fails to compute the control strategies within 168 hours (one week). The GATTACA framework identifies 165 PA states and returns control strategies of average length of 7.1 for the target configuration $x_3 = 1$, where $x_3$ is one of the nodes; see Table~\ref{tab:results-ATGC}. This relatively long control strategy on average can be explained by the fact that a~network of highly interconnected nodes may require a~simultaneous perturbation of a~larger number of nodes than in the case of a~loosely connected network. Similarly to the two cases of the original T-diff model in Table~\ref{tab:tdiff}, since we limit the number of simultaneous perturbations, the GATTACA framework finds longer control strategies that traverse a~number of intermediate PA states. Training of the GATTACA framework on the 200-node network required approximately 2.5h (A100) and 0.6M steps.

\section{Conclusions}
\label{sec:discussions}

We formulate the attractor-target control problem in the context of cellular reprogramming --~a~novel control problem for biological networks modelled using the BN framework under the asynchronous update mode. To provide a~scalable solution, we develop and implement the GATTACA framework. Our framework can handle models of sizes that are prohibitive for existing state-of-the-art control algorithms.

To facilitate scalability, the GATTACA framework is based on DRL. Importantly, to leverage the known structure of a~biological system and to encode the network dynamics into a~latent representation, we incorporate graph neural networks with graph convolution operations into the artificial neural network approximator for the action-value function learned by the ATGC DRL agent. The available knowledge of the structure of a~BN model is explicitly incorporated into the ATGC by defining the GCN architecture in accordance with the structure graph of the BN. Through interactions with the BN behaviour determined by the Boolean functions of the BN, the BN model dynamics is implicitly encoded into the ATGC via training. Moreover, the GCN creates an~embedding of a~discrete graph into a~continuous space, which enhances the training of the DRL agent.

Moreover, scalability is achieved by considering the previously introduced concept of a~pseudo-attractor and by exploiting iPASIP, which we develop by improving previously proposed approach for effective identification of PA states in large BN models. We demonstrate the effectiveness of iPASIP on a~number of BN models of real-world biological networks of different sizes by comparing the identified PA states with true attractors obtained with an~exact state-of-the-art algorithm implemented in the CABEAN software tool. We show that iPASIP is capable of correctly identifying the representative attractor states of all attractors in most of the cases. It also manages to detect attractor states in BN models for which CABEAN fails to compute the attractors. Furthermore, iPASIP proves its scalability and effectiveness as part of the GATTACA framework, where the determined PA states facilitate the computation of control strategies for large BN models of various real-world biological networks. 

The proposed GATTACA framework offers a~scalable and effective solution to the attractor-target control problem, as demonstrated by the results of the conducted computational experiments on a~number of large real-world biological networks from the literature. Given the source PA state and target configuration of genes, the ATGC of the GATTACA framework finds proper control strategies that drive the network from the source PA to the target configuration by taking actions (intervening) only in intermediate PA states that correspond to phenotypical cellular states that can be observed in the lab. Our framework is capable of finding optimal or close to optimal control strategies, where in the latter case the lengths in most of the cases differ on average from the optimal strategies by at most one intervention. Moreover, the GATTACA framework is capable of identifying effective control strategies of reasonable lengths in BN models on which CABEAN fails. These models are large in terms of number of nodes and often characterised by many (pseudo-)attractor states. An~additional evaluation conducted on a~challenging network which structure graph is a~single large SCC of 200 nodes, the analysis of which is beyond the capabilities of state-of-the-art exact solutions, further demonstrates the unprecedented scalability potential of the GATTACA framework in controlling large Boolean models of complex biological networks.

\iftrue
\section*{Acknowledgments}

This research was funded in whole or in part by the National Science Centre, Poland under the OPUS call in the Weave programme, grant number 2023/51/I/ST6/02864.

For the purpose of Open Access, the author has applied a~CC-BY public copyright licence to any Author Accepted Manuscript (AAM) version arising from this submission.
\fi

\bibliographystyle{IEEEtran}
\bibliography{mybib}

%\printbibliography

\appendix

\renewcommand\thefigure{\thesection.\arabic{figure}}
\setcounter{figure}{0}
\renewcommand{\thetable}{A.\arabic{table}}
\setcounter{table}{0}

\subsection{Example: Boolean network}
\label{app:bn_example}

Let us consider a~Boolean Network $\textsf{BN} = (\mathbf{x},\mathbf{f})$, where $\mathbf{x}=(x_1,x_2, x_3)$ and $\mathbf{f} = (f_1, f_2, f_3)$ is defined as:
$$\begin{cases}
f_1 = x_1, \\
f_2 = x_1 \lor x_3, \\
f_3 = x_2 \land x_3.
\end{cases}$$
Its state transition graph under the asynchronous update mode is shown in Figure~\ref{fig:stg}. In this network, $x_1$ is an~\emph{input node}, meaning that its state remains constant during the evolution of the system. Let us consider a~target configuration defined by $x_2 = 0$; that is, we want the network to settle into an attractor where $x_2$ set to 0, while the states of the other nodes are not constrained.

\begin{figure}[ht]
\centering
% \begin{tikzpicture}[x=3.5cm,y=3.5cm] % change these values to adjust the size of a figure
%   \tikzset{
%     e4c node/.style={circle,draw,minimum size=1.2cm,inner sep=0}, 
%     e4c edge/.style={sloped,above,font=\footnotesize}
%   }
%   \node[fill=blue!30][e4c node, text=red] (0) at (0.0, 2.3) {(0,0,0)}; 
%   \node[e4c node] (1) at (0.0, 3.0) {(0,0,1)}; 
%   \node[e4c node] (2) at (1.0, 2.3) {(0,1,0)}; 
%   \node[fill=blue!30][e4c node] (3) at (1.0, 3.0) {(0,1,1)}; 
%   \node[fill=blue!30][e4c node, text=red] (4) at (0.0, 4.4) {(1,0,0)}; 
%   \node[e4c node] (5) at (0.0, 3.7) {(1,0,1)}; 
%   \node[fill=blue!30][e4c node] (6) at (1.0, 4.4) {(1,1,0)}; 
%   \node[fill=blue!30][e4c node] (7) at (1.0, 3.7) {(1,1,1)}; 

%   \path[->,draw,thick]
%   (1) edge[e4c edge] (0)
%   (2) edge[e4c edge] (0)
%   (1) edge[e4c edge] (3)
%   (4) edge[e4c edge, bend left=20] (6)
%   (6) edge[e4c edge, bend left=20] (4)
%   (5) edge[e4c edge] (4)
%   (5) edge[e4c edge] (7)
%   ;
% \end{tikzpicture}

\begin{tikzpicture}[x=4cm,y=4cm] % change these values to adjust the size of a figure
  \tikzset{     
    e4c node/.style={circle,draw,minimum size=0.7cm,inner sep=0}, 
    e4c edge/.style={sloped,above,font=\footnotesize}
  }
  \node[fill=blue!30][e4c node, text=red] (0) at (-0.5, 2) {\footnotesize{(0,0,0)}}; 
  \node[e4c node] (1) at (-0.5, 2.35) {\footnotesize{(0,0,1)}}; 
  \node[e4c node] (2) at (0, 2) {\footnotesize{(0,1,0)}}; 
  \node[fill=blue!30][e4c node] (3) at (0, 2.35) {\footnotesize{(0,1,1)}}; 
  \node[fill=blue!30][e4c node, text=red] (4) at (0.5, 2.35) {\footnotesize{(1,0,0)}}; 
  \node[e4c node] (5) at (0.5, 2) {\footnotesize{(1,0,1)}}; 
  \node[fill=blue!30][e4c node] (6) at (1, 2.35) {\footnotesize{(1,1,0)}};
  \node[fill=blue!30][e4c node] (7) at (1, 2) {\footnotesize{(1,1,1)}};

  \path[->,draw,thick]
  (1) edge[e4c edge] (0)
  (2) edge[e4c edge] (0)
  (1) edge[e4c edge] (3)
  (4) edge[e4c edge, bend left=20] (6)
  (6) edge[e4c edge, bend left=20] (4)
  (5) edge[e4c edge] (4)
  (5) edge[e4c edge] (7)
  ;
\end{tikzpicture}

\caption{State transition graph with highlighted attractor states of the Boolean network in under the asynchronous update mode.}
\label{fig:stg}
\end{figure}

If the input node $x_1 = 0$, then the target configuration is realised only by the attractor state $(0, 0, 0)$, which can be reached from the other attractor through a~single perturbation:
$$(0, 1, 1) \xrightarrow{i=3} (0, 1, 0) \xrightarrow{\text{evolution}} (0, 0, 0),$$
where $\xrightarrow{i=3}$ denotes a~change in state caused by flipping the value of node $x_3$, and $\xrightarrow{\text{evolution}}$ represents a~sequence of transitions consistent with the natural dynamics of the network.

If $x_1 = 1$, there is only one attractor state that realises the target configuration, namely $(1, 0, 0)$. This state is reached through natural evolution from $(1, 1, 0)$, while a~single perturbation of the input node is required to reach it from $(1, 1, 1)$:
$$(1,1,1) \xrightarrow{i=1} (1, 1, 0) \xrightarrow{\text{evolution}} (1, 0, 0).$$

\subsection{Example: Attractor-target control with the switch of environmental conditions}
\label{app:atc_example}

We consider a~BN with 5 nodes, i.e. $x_1, x_2, x_3, x_4, x_5$. The nodes $x_1$ and $x_2$ are designated as input nodes and the control objective is to achieve the target configuration $x_5 = 1$. Let us assume that the network is as follows.
\begin{itemize}
  \item For \( x_1 = 0, x_2 = * \), the network has two fixed-point attractors:
  \[
    \{(0, 0, 0, 0, 0)\}, \quad \{(0, 0, 1, 1, 0)\}.
  \]
  In both cases $x_5 = 0$, so there is no attractor aligned with the target configuration.
    
  \item For \( x_1 = 1, x_2 = 1 \), the attractors are:
    \[
    \{(1, 1, 1, 0, 1)\}, \quad \{(1, 1, 0, 1, 0)\}.
    \]
    
    \item For \( x_1 = 1, x_2 = 0 \), the attractor is:
    \[
    \{(1, 0, 0, 0, 1)\}.
    \]
\end{itemize}
Thus, there are three source attractors and two target attractors.

We consider $\{(0,0,0,0,0)\}$ to be the source attractor. Without perturbations of the input nodes $x_1$ and $x_2$, the system remains trapped in the configuration $x_5 = 0$. Therefore, perturbations of input nodes are necessary to change the environmental condition. There are two options: set $x_1 = 1, x_2 = 1$ or $x_1 = 1, x_2 = 0$. We consider both.
\begin{itemize}
  \item Suppose that the perturbation to state $(1, 1, 0, 0, 0)$ finally leads to the attractor:
    \[
    \{(1, 1, 0, 1, 0)\}.
    \]
  Since $x_5 = 0$, an~additional intervention on $x_3$ is required to reach the target attractor, which is a~CABEAN control path of length 1. Therefore, the total control strategy length in this case is 2.

  \item The perturbation to state $(1, 0, 0, 0, 0)$ leads directly to the attractor:
  \[
    \{(1, 0, 0, 0, 1)\},
  \]
  which is a~target attractor. Here, the total control strategy length is 1.
\end{itemize}
Consequently, the length of the reference control strategy obtained with CABEAN is 1 for this example.

\subsection{Choosing iPASIP hyperparameters}
\label{app:iPASIP_hyperparams}

The iPASIP hyperparameters should be set on a~case-by-case basis in accordance with the available prior knowledge on the system under study, i.e. expected number and size of attractors. Additionally, for the setting of the $k_1$ hyperparameter (and $k_2$ by analogy) the information on the upper bounds on the size of the associated pseudo-attractors can be exploited as provided by the following theorem, which was originally proved in~\cite{ours_tcs}.

\begin{theorem}
For any attractor $\mathcal{A}$ of a~BN under the asynchronous update mode, the size of the associated pseudo-attractor $\mathcal{PA}$ found by Step~I-1 of the pseudo-attractor identification procedure with $k_1=k\%$ identification threshold is exactly upper bounded by
\[
|\mathcal{PA}| \leq 
\begin{cases}
\frac{100}{k}-1, & 100 \text{ mod } k = 0 \text{ and } |\mathcal{A}| > \frac{100}{k} \\
\floor{\frac{100}{k}}, & \text{otherwise}.
\end{cases}
\]
\label{th:p-a_size_bounds}
\end{theorem}

\iftrue
\begin{proof}
Since all states of $A$ are positive recurrent in the Markov chain underlying the BN restricted to $A$, it holds that $\mathbb{P}_A(s) > 0$ for all $s \in A$. We start with the first case, where $100 \text{ mod } k = 0$ and $|A| > \frac{100}{k}$. A~state $s$ is identified as a~PA~state by Step~I if $\mathbb{P}_A(s) \geq \frac{k}{100}$. A~pseudo-attractor associated with $A$ will be of maximum size whenever the stationary distribution maximises the number of states with $\frac{k}{100}$ probability. The maximum must be less than $\frac{100}{k}$. To show this, let $M$ denote the subset of states of $A$ with stationary distribution probability of $\frac{k}{100}$ and let assume that $|M| \geq \frac{100}{k}$. Then, $\sum_{s \in A}\mathbb{P}_A(s) =  \sum_{s \in M}\mathbb{P}_A(s) + \sum_{s \in A\setminus M} \mathbb{P}_A(s) \geq \frac{k}{100} \cdot \frac{100}{k} + \sum_{s \in A\setminus M} \mathbb{P}_A(s) > 1$, where the last inequality follows from the fact that all the states in $A\setminus M$ have non-zero stationary probabilities. Therefore, $\mathbb{P}_A$ cannot be a~probability distribution. Hence, $|M| \leq \frac{100}{k} - 1$. The upper bound is reached, for example, by any~stationary distribution $\mathbb{P}_A$ such that the for the $\frac{100}{k} - 1$ states in $M$ the probabilities are equal to $\frac{k}{100}$ and $\sum_{s \in A\setminus M} \mathbb{P}_A(s) = \frac{k}{100}$. This concludes the first case. 

In the other case, let $\epsilon := \frac{100}{k} - \floor{\frac{100}{k}}$. If $|A| \leq \floor{\frac{100}{k}}$, then the inequality holds since $PA \subseteq A$ by definition. The upper bound is reached if $|A| = \floor{\frac{100}{k}}$ and, for example, the stationary distribution probabilities for $|A|-1$ states in $A$ are equal to $\frac{k}{100}$ and for the remaining state the probability is $\frac{k}{100}\cdot\left(1+\epsilon\right)$. If $|A| > \floor{\frac{100}{k}}$ in the second case, then $\epsilon > 0$, i.e. $\epsilon \in (0,1)$. The maximum possible number of states of $A$ with stationary probability equal $\frac{k}{100}$ must be less than $\floor{\frac{100}{k}} + 1$. To see this, let $M$ be the set of such states of $A$ and let assume that $|M| \geq \floor{\frac{100}{k}} + 1$. Then, $\sum_{s \in A}\mathbb{P}_A(s) > \frac{k}{100}\cdot \left(\floor{\frac{100}{k}} + 1\right) > \frac{k}{100}\cdot\left(\floor{\frac{100}{k}} + \epsilon\right) = 1$. This contradicts the fact that $\mathbb{P}_A$ is a~probability distribution on $A$. Finally, we proceed to show that the upper bound can be reached also in this scenario. This is the case, for example, for a~stationary probability distribution $\mathbb{P}_A$ on $A$ where $\floor{\frac{100}{k}}$ states of set $M$ have probabilities $\frac{k}{100}$ and all the states in $A\setminus M$ have non-zero probabilities such that $\sum_{s \in A\setminus M} \mathbb{P}_A(s) = \epsilon$. This is possible since $\epsilon > 0$.
\end{proof}
\fi

\end{document}